\documentclass[preprint]{elsarticle}
\makeatletter
\def\ps@pprintTitle{%
     \let\@oddhead\@empty
     \let\@evenhead\@empty
     \def\@oddfoot
       {\hbox to \textwidth%
        {\ifnopreprintline\relax\else
        \@myfooterfont%
         \ifx\@elsarticlemyfooteralign\@elsarticlemyfooteraligncenter%
           \hfil\@elsarticlemyfooter\hfil%
         \else%
         \ifx\@elsarticlemyfooteralign\@elsarticlemyfooteralignleft%
           \@elsarticlemyfooter\hfill{}%
         \else%
         \ifx\@elsarticlemyfooteralign\@elsarticlemyfooteralignright%
           {}\hfill\@elsarticlemyfooter%
         \else%
               Preprint submitted to \ifx\@journal\@empty%
                 Elsevier%
            \else\@journal\fi\hfill\@date\fi%
         \fi%
         \fi%
         \fi%
         }%
       }%
     \let\@evenfoot\@oddfoot}
\makeatother

\usepackage[hyphens]{url}
\usepackage{hyperref}

\usepackage{amssymb}
\usepackage[ruled,vlined]{algorithm2e}
\usepackage{caption}
\usepackage{subcaption}
\usepackage{makecell}
\usepackage{longtable}
\usepackage{svg}
\usepackage{amsmath}
\usepackage{graphicx}
\usepackage{geometry}
\usepackage{xcolor}

\usepackage[textsize=tiny]{todonotes}
\hypersetup{pdfauthor=author}

\usepackage{float}
\restylefloat{table}

\usepackage{array}
\newcolumntype{P}[1]{>{\centering\arraybackslash}p{#1}}
\newcolumntype{M}[1]{>{\centering\arraybackslash}m{#1}}

\usepackage{etoolbox}
\apptocmd{\thebibliography}{\raggedright}{}{}

\journal{Arxiv.org}

\begin{document}

\begin{frontmatter}

\title{Natural Language Processing for Systems Engineering: Automatic generation of Systems Modeling Language Diagrams}

\author[firstaddress]{Shaohong Zhong}
\address[firstaddress]{Department of Engineering Science, University of Oxford, Parks Road, Oxford, OX1 3PJ, United Kingdom}

\author[secondaddress]{Andrea Scarinci}
\address[secondaddress]{Uncertainty Quantification Laboratory, Department of Aeronautics and Astronautics, Massachusetts Institute of Technology, Cambridge, MA 02139, United States}
            
\author[firstaddress,thirdaddress]{Alice Cicirello\corref{mycorrespondingauthor}}
\ead{a.cicirello@tudelft.nl}
\cortext[mycorrespondingauthor]{Corresponding author}
\address[thirdaddress]{Depeartment of Engineering Structures, Delft University of Technology, Stevinweg 1, Delft, 2628 CN, The Netherlands}


            

\begin{abstract}

The design of complex engineering systems is an often long and articulated process that highly relies on engineers' expertise and professional judgment. As such, the typical pitfalls of activities involving the human factor often manifest themselves in terms of lack of completeness or exhaustiveness of the analysis, inconsistencies across design choices or documentation, as well as an implicit degree of subjectivity. An approach is proposed to assist systems engineers in the automatic generation of systems diagrams from unstructured natural language text. Natural Language Processing (NLP) techniques are used to extract entities and their relationships from textual resources (e.g., specifications, manuals, technical reports, maintenance reports) available within an organisation, and convert them into Systems Modelling Language (SysML) diagrams, with particular focus on structure and requirement diagrams. The intention is to provide the users with a more standardised, comprehensive and automated starting point onto which subsequently refine and adapt the diagrams according to their needs. The proposed approach is flexible and open-domain. It consists of six steps which leverage open-access tools, and it leads to an automatic generation of SysML diagrams without intermediate modelling requirement, but through the specification of a set of parameters by the user. The applicability and benefits of the proposed approach are shown through six case studies having different textual sources as inputs, and benchmarked against manually defined diagram elements. 

\end{abstract}

\begin{keyword}
SysML diagram \sep NLP \sep Structure diagram \sep Requirement diagram \sep text-to-diagram \sep automated diagrams from text

\end{keyword}

\end{frontmatter}


\section{Introduction} 

Systems engineering is a crucial process in the design and management of complex systems \cite{kossiakoff2011systems,friedenthal14practical,hart2015introduction,delligatti2013sysml,huang2007system}. An integral part of designing and architecting engineering systems involves producing formal documentation to both record and support the development process (e.g. systems specifications \cite{dori2004smart}, codified system diagrams etc.). \textcolor{black}{An open debate within the systems engineering community exists about the notion of \textit{completeness} of specifications (requirements in particular), and means to ``check'' for it.} \textcolor{black}{In this paper, the authors are not concerned with establishing a new paradigm for completeness verification. Instead, the authors aim at facilitating a crucial step in achieving such a goal: the processing of the large amounts of documentation and engineering knowledge from which specifications are drawn.} Maintaining document consistency and/or completeness while gathering the necessary information to specify a system's behaviour is both challenging and crucial to avoid unexpected consequences that may result in uncontrolled hazards during the system's operations \cite{sawyer2005shallow,arellano2015frameworks}. Errors and omissions in the initial design stages in particular can lead to costly product modifications after manufacturing \cite{arellano2015frameworks}.

\textcolor{black}{Typically, the synthesis of available knowledge into appropriate engineering documentation and formats (e.g. requirements, diagrams) is a manual process. With this comes} a certain degree of subjectivity and arbitrariness. \textcolor{black}{In this paper, the authors propose a way of automating this manual process by using Natural Language Processing (NLP) techniques: on the one hand, having machine learning algorithms replacing manual work decreases costs and processing times; on the other hand, systematizing the gathering of information and leaving the engineers with a purely supervisory task could provide for better coverage and consistency.}

\textcolor{black}{As a testbed for the concept just described, the authors focus on automating the generation of a very commonly used tool in systems design: SysML diagrams.}  SysML is a graphical modelling language that is a subset of the Unified Modelling Language (UML) with extensions catered to systems engineering \cite{friedenthal14practical,delligatti2013sysml,huang2007system,omg19omg,friedenthal08omg,hause06thesysml}. It is able to support the specification, analysis, and design of complex systems such as hardware, software, and facilities \cite{friedenthal14practical}. SysML has also gained importance in recent years as a critical enabler of Model-based Systems Engineering \cite{friedenthal14practical,hart2015introduction}. Like many other graphical modelling languages it still requires significant manual input \cite{friedenthal14practical,hart2015introduction,delligatti2013sysml}. To the best of the authors’ knowledge, this is the first study that focuses on the automatic generation of SysML diagrams from unstructured natural language text.

The main contributions of the paper are as follows:

\begin{itemize}
\item The design of a versatile automated approach to generate SysML diagrams from natural language text documents;
\item The combined use of NLP techniques and the lexical databases to facilitate and augment the generation of SysML models;
\item The mapping of textual entities (phrases and relationships) to SysML model elements using NLP techniques and heuristic rules.
\end{itemize}

The rest of this paper is structured as follows: a detailed overview of the current state-of-the-art  NLP-based strategies for automating systems engineering processes is provided in Section~\ref{section:litreview}. The proposed approach is presented in Section~\ref{section:methods}, and the steps for its implementation are summarised in Section~\ref{section:implementation}. The case studies and experimental procedures are then described in Section~\ref{section:exp}. In Section~\ref{section:results}, the results are discussed.

\section{Review of state-of-the-art NLP-based techniques for automating \textcolor{black}{System Engineering diagram generation}}
\label{section:litreview}
\textcolor{black}{To the best of the authors' knowledge, the automatic generation of requirement diagrams and structure diagrams has not been addressed in the literature yet. Therefore, this section focuses on reviewing the state-of-the-art of three key topics: (i) NLP in System Engineering for the generation of system engineering models; (ii) Automatic generation of Unified Modelling Language (UML) diagrams; (iii) Ontology learning for automatic extraction of concepts, relations, attributes, and hierarchies from text. }

\subsection{NLP in Systems Engineering \textcolor{black}{for the generation of system engineering models}}
NLP techniques have been used in requirements engineering, where the majority of requirements documents and their sources, such as user reviews, are written in natural language \cite{zhao2021natural,bakar2016extracting,robeer2016automated,johann2017safe}. Past studies have used NLP for requirement elicitation \cite{abad2019supporting}, requirement enrichment \cite{korner2010semantic}, requirements tracing \cite{hayes2003improving}, requirement classification \cite{casamayor2010identification}, requirement improvement \cite{rodriguez2019efficient,ferrari2013mining}, identifying the domain of requirement \cite{thakur2016identifying}, ambiguity detection \cite{ferrari2019nlp}, generating test cases and use cases \cite{carvalho2013test,silva2016test,tiwari2019approach}, and detection of low-quality requirements \cite{ferrari2018detecting}. A more detailed review is done in \cite{zhao2021natural}, which classified the function of NLP in requirement engineering as detection, extraction, classification, modelling, tracing and relating, and search and retrieval. Of specific interest here are studies that have attempted to extract requirements and construct models from natural language text, \textcolor{black}{which are briefly reviewed in what follows}. 

\textcolor{black}{Recent works,  used NLP for mining concepts from a mix of textual requirement assets such as documentation and manuals ~\cite{loughran2006from}, product brochures ~\cite{ferrari2013mining}, online reviews~\cite{bakar2016extracting}, app descriptions and reviews \cite{johann2017safe}. However, these works rely on the user to specify relations between identified concepts and to manually construct the system engineering models~\cite{ferrari2013mining}, even when the goal is to generate structure features diagrams~\cite{ferrari2013mining}.  }

\textcolor{black}{The state-of-the-art approaches for the generation of systems engineering models from natural language text inputs are based on the specification of a set of heuristics rules and by predefining either a set of association phrases ~\cite{sreekumar2018extracting}, or a mix of manually specified sentence patterns \cite{al2009natural, robeer2016automated}, or a predefined list of goal-specific keywords and syntactic patterns \cite{casagrande2014nlp}.}  Other studies have also used NLP techniques such as dependency parsing and coreference resolution to identify type dependencies. The hierarchy was then identified using predefined rules or classifiers \cite{thayasivam2011automatically,nguyen2015rule}. However, these approaches are limited by their reliance on heuristics rules \cite{thayasivam2011automatically}. Additionally, they also tended to put strict constraints on the forms of inputs, either specifying complex syntactic rules or restricting inputs to be of a certain format, for example, requirement specification documents \cite{robeer2016automated,sreekumar2018extracting,al2009natural,casagrande2014nlp}. Such constraints limit the flexibility of the approaches to adapt to the variety of textual assets available for a systems engineer and require a significant amount of prior work to produce structured requirement documents for processing. 

\subsection{Modelling in SysML and UML}
Unlike requirement diagrams, which are unique to the SysML profile, structure diagrams in SysML bear a close resemblance to UML diagrams such as class diagrams and composite structure diagrams \cite{friedenthal14practical,hause06thesysml,omg19omg}. Thus, past studies aimed at generating UML diagrams from natural language text are also reviewed. 

In~\cite{chen2009automatic}, the authors proposed a method to automatically generate UML class diagrams from natural language requirements by employing a Recursive Object Model (ROM) as the intermediate step. The ROM is a graphical language that treats each word and each sentence as objects and assigns semantic relations from a predefined set to these objects using lexical and syntactic rules~\cite{zeng2008recursive}. The ROM model was then traversed and converted to a class diagram using the noun objects and their relations~\cite{zeng2008recursive}. Additionally, in \cite{afreen2011generating}, the authors used PoS tagging and semantic role labelling combined with manually defined rules to generate class diagrams from requirement specifications. However, similar to past studies on generating requirement models, these studies rely on using structure requirement documents of specific formats as inputs and are thus limited in their flexibility \cite{chen2009automatic,afreen2011generating}. In contrast, in \cite{deeptimahanti2011semiautomatic}, the authors proposed a method to automatically generate UML class diagrams from natural language stakeholder requests. They used a set of predefined syntactic rules to decompose the inputs into simple sentences, then parsed the resulting sentences and identified classes and relations using a set of heuristic rules \cite{deeptimahanti2011semiautomatic}. However, the reliance on predefined rules on syntactic features to classify relations also limit the wider applicability of this approach. 

\subsection{Ontology learning}
\textcolor{black}{Ontology learning refers to the automatic extraction of ‘concepts, relations, attributes, and hierarchies’ from text \cite{gruber1995toward,asim2018survey}. State-of-the-art methods to automatically extract ontology models from natural language text include Text2Onto~\cite{cimiano2005text2onto}, OntoGain~\cite{drymonas2010unsupervised}, a graph-based algorithm~\cite{velardi2013ontolearn}, and CRCTOL ~\cite{jiang2010crctol}.  Ontology Learning techniques could be used to enable the automatic generation of SysML diagrams. This is because, similarly to Ontology Learning, System Engineering modelling approaches employ an object-centred representation of the target domain and use similar constructs such as classes/blocks and relations \cite{mejhedmkhinini2020combining}. However, as highlighted in~\cite{mejhedmkhinini2020combining}, the techniques in ontology generation need to be adapted to suit systems engineering. For example, the primary source for generating an informative diagram is a limited corpus, instead of relying on the web. Moreover, the extracted structured information has to be classified into different types of SysML diagrams \cite{omg19omg}.}

\subsection{\textcolor{black}{An approach to overcome the limitations of the state-of-the-art}}
\label{subsection:researchq}
\textcolor{black}{Based on the above review, it has been identified the need for the development of an automatic approach to generate SysML diagrams from unstructured natural language text. The proposed approach is based on natural language processing techniques and the semantic web, and goes beyond the state-of-the-art by tackling the following research questions}:

\begin{itemize}
\item \textcolor{black}{How to extract information for developing system engineering models without specifying restrictive syntactic rules valid only for a specific domain?}
\item \textcolor{black}{How to identify hierarchies with minimal reliance on heuristic rules and predefined patterns?}
\item \textcolor{black}{How to automate the generation of different types of SysML diagrams with no human intervention or intermediate modelling from a limited corpus of textual resources?}
\end{itemize}

\section{Automatic generation of Systems Modeling Language diagrams}
\label{section:methods}

The proposed approach focuses on automatically generating a subset of structure and requirements diagrams in SysML from a corpus of natural language text documents. Specifically, this paper focuses on Block Definition Diagrams (BDD), Internal Block Diagrams (IBD), and Requirement Diagrams (REQ) \cite{omg19omg}. The basic elements of these three diagrams include blocks and their relationships. A block is an elemental modelling construct in SysML that represents both real entities, such as physical objects, and abstract entities, such as concepts~\cite{friedenthal14practical}. Relationships between blocks in SysML can be further classified into categories such as association or generalisation for BDD, and trace or containment for REQ \cite{friedenthal14practical,omg19omg,friedenthal08omg,hause06thesysml}. 

To generate these diagram elements, a parallel architecture is proposed to extract two kinds of textual entities from natural language text: 1) key phrases, where a key phrase is defined as a list of one or more words that represent an important entity described in the text. 2) key relationships between the key phrases, where a key relationship is defined as one that links two key phrases. The key phrases and key relationships are used to generate blocks and the relationships between blocks, respectively. The generated blocks and their relationships are then organised into desired SysML diagrams. 

The procedures for automatic diagram generation are summarised in Figure~\ref{fig:flowchart} and it consists of six steps:

\begin{enumerate}
  \item  The first step is the manual selection and upload of a corpus of text documents (Section~\ref{subsection:rawtextselection}).
  
  \item The raw texts selected are then used as inputs for the key noun extraction (Section~\ref{subsection:nounextraction}).
  
  \item The raw texts selected are used also as inputs relation extraction (Section~\ref{subsection:relationshipextraction}). 
  
  \item Steps 2 and 3 result in a collection of key nouns and a collection of relations, respectively. The two collections are then used to generate the list of key phrases and key relations (Section~\ref{subsection:keyselection}).
  
  \item The list of key phrases and key relations are subsequently used to generate corresponding SysML model elements and augmented according to the required diagram type (Section~\ref{subsection:mappingandaugmentation}).
  
  \item Finally, the generated SysML model elements are then organised and plotted to the corresponding SysML profile (Section~\ref{subsection:diagramgeneration}). 
\end{enumerate}

An illustrative example is provided in Figure~\ref{fig:flowcharteg}. Each step is discussed in detail in what follows. 
\newline
\newline
\newline

\begin{figure}[ht!]
    \centering
    \begin{subfigure}[b]{0.6\columnwidth}
    \centering
    \includegraphics[width=0.45\columnwidth]{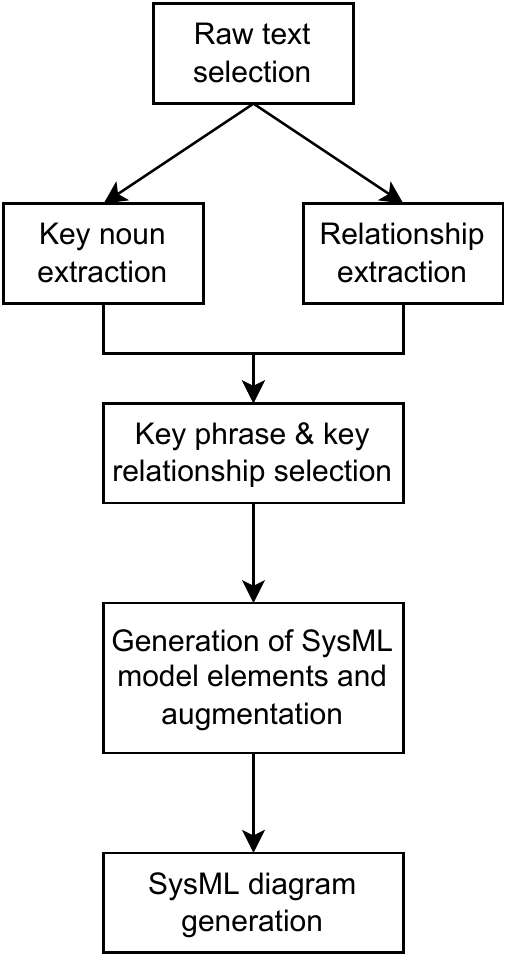}
    \caption{Steps for automatic SysML diagram generation
    }
    \label{fig:flowchart}
    \end{subfigure}
\end{figure}

\break
\break
\break

\begin{figure}[ht!]\ContinuedFloat
    
    \begin{subfigure}[b]{\columnwidth}
    \centering
    \includegraphics[width=0.6\columnwidth]{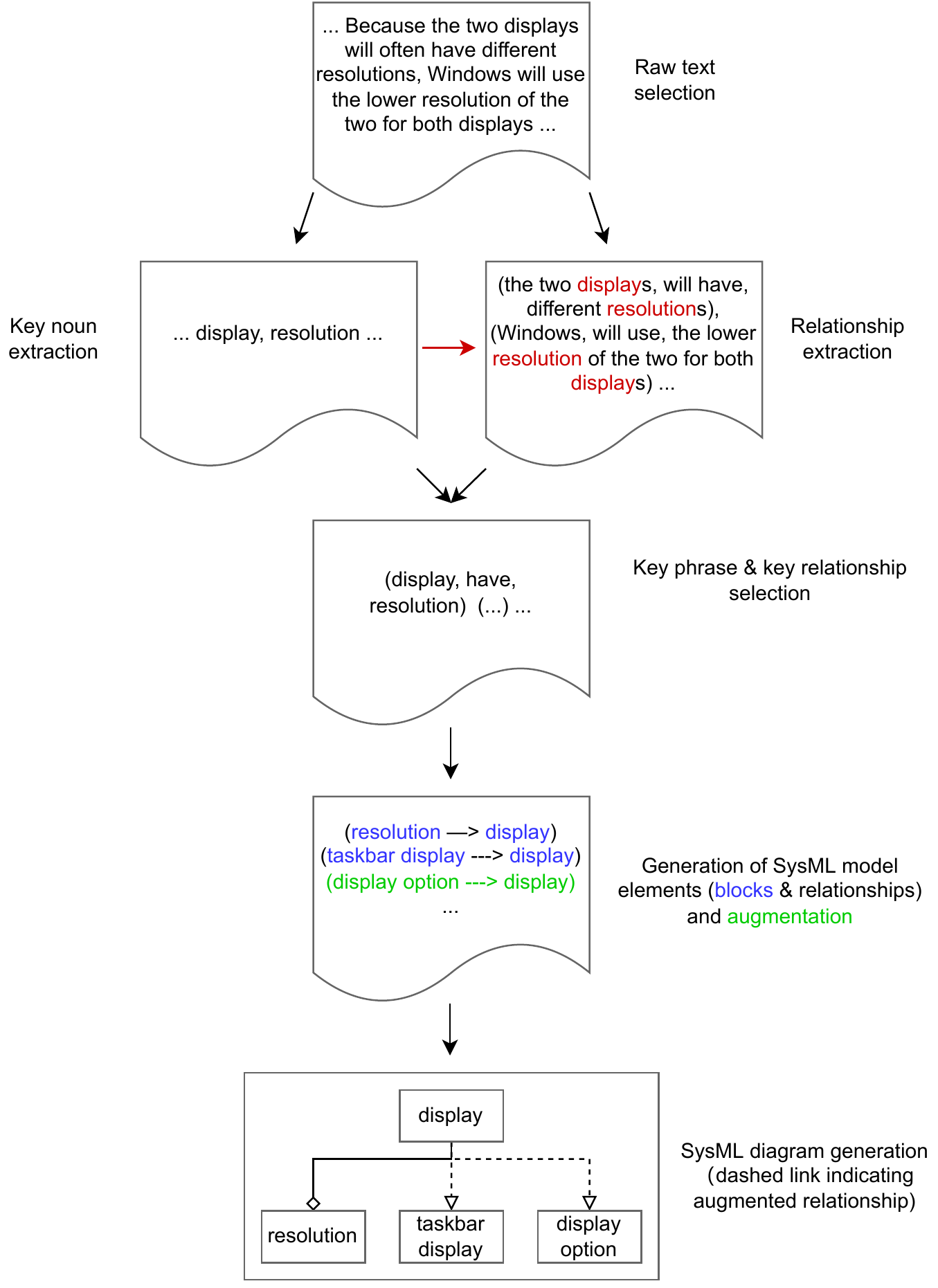}
    \caption{Illustration of automated diagram generation steps using an excerpt from a Windows manual as example
    }
    \label{fig:flowcharteg}
    \end{subfigure}
\caption{Procedures and examples for automatic SysML diagram generation}
\end{figure}

\subsection{Raw texts selection}
\label{subsection:rawtextselection}
The algorithm takes as input a manually selected corpus of text documents. These documents do not need to be structured or follow a predefined writing style like requirement documents, and the proposed approach can be applied to documents typically available to systems engineers, for example manuals \cite{friedenthal14practical,delligatti2013sysml}. \textcolor{black}{Additionally, the corpus should be of sufficient size. Providing a universal standard for the sufficient size of textual resources, covering different domains and writing styles, is a challenging task that goes beyond the present work. Nonetheless, based on the experiments carried out by the authors (with the most salient ones reported in Section~\ref{section:exp}), it is recommended that for the successful generation of diagrams, the document count should be larger than 100 and the average word count per document should be larger than 500.  A corpus can also be obtained by splitting one large document, for example one manual book, into chapters and sections, as illustrated in the case study on UK government reports and a Windows manual in Section~\ref{section:exp}}.

\subsection{Key nouns extraction}
\label{subsection:nounextraction}
This step of the approach aims to select key nouns, which are nouns that represent important entities and that can serve as constituents of the key phrases. The key nouns are automatically extracted from the corpus by employing preprocessing and term frequency-inverse document frequency (tf-idf) techniques \cite{jurafsky2009speech}.

\subsubsection{Preprocessing}
\label{section:preprocessing}
Preprocessing is needed to remove grammatical features in the text to support noun extraction \cite{jurafsky2009speech}. In this study, a combination of preprocessing methods is applied to the text inputs: 
 
\textit{Tokenisation}: Tokenisation breaks down large textual content, such as paragraphs or documents, into smaller chunks \cite{jurafsky2009speech}. In this study, the Natural Language Toolkit (NLTK) is used to tokenise each input text document into individual sentences and then tokenise each sentence into individual words. \textcolor{black}{In particular, the adapted Punkt sentence segmenter for sentence tokenisation \cite{kiss2006unsupervised} and the Penn Treebank Tokeniser for word tokenisation \cite{bird2009natural} available in NLTK are used. In this work, only one-word unigrams are considered.} 


\textit{Part-of-speech (PoS) tagging}: The NLTK package is used to classify each word of a sentence into different lexical categories. The PoS tagger assigns a PoS tag to each word based on its context sentence \cite{jurafsky2009speech}. \textcolor{black}{The Greedy Averaged Perceptron tagger and the Penn Treebank tagset are used \cite{bird2009natural}.} Examples of PoS tags include present tense verbs (‘VBP’) and adjectives (‘JJ’). After the PoS tag for each word is assigned, the nouns are used for further processing and words of other types are removed. 

\textit{Lemmatisation}: The individual nouns are further \textcolor{black}{converted to lower case and} reduced to their root forms through the NLTK WordNet Lemmatiser, e.g. from ‘sensors’ to ‘sensor’. The WordNet Lemmatiser employs a combination of predefined rules and dictionary search in lemmatisation \cite{bird2009natural}.
 
\textit{Stop word removal}: As the last step, all stop words that are still in the text are removed, using the NLTK list of English stop words as a reference \cite{bird2009natural}. 
 
\subsubsection{Evaluation of term frequency-inverse document frequency}
\label{section:evaluationoftfidf}
The term frequency-inverse document frequency (tf-idf) metric is employed to evaluate the domain relevance of each noun after preprocessing \cite{jurafsky2009speech,luhn1957statistical,jones1972statistical}. Note that after the preprocessing steps, the documents contain only nouns. Tf-idf treats each word as  unigram, and word orders or document orders are not considered. The tf-idf weighting $w$ of a word $t$, in a document $d$ belonging to a corpus of documents $c$, is a value computed as \cite{jurafsky2009speech,luhn1957statistical,jones1972statistical}

\begin{equation}
w_{t,d} = \textup{tf}_{t,d} \times \textup{idf}_{t,c}                                                         
\end{equation}

$\textup{tf}_{t,d}$ is given by

\begin{equation}
\textup{tf}_{t,d} = \textup{log}_{10}(count_{t,d}+1)                                                        
\end{equation}
where $count_{t,d}$ is the total count of a word $t$ in a given document $d$, and the inverse document frequency $\textup{idf}_{t,c}$ is given by \cite{jurafsky2009speech,jones1972statistical}
\begin{equation}
\textup{idf}_{t,c}  = \textup{log}_{10}( N_c / (1 + \textup{df}_{t,c}) ) + 1                          
\end{equation}

where $N_c$ is the total number of documents in the corpus $c$, and the document frequency $\textup{df}_{t,c}$ is the number of documents in the corpus $c$ that contain the word $t$ \cite{jurafsky2009speech}. In this study, the tf-idf weighting is normalised to the range $0 \leq w_{t,d} \leq 1$ by dividing by the largest tf-idf weighting in a document. The inclusion of inverse document frequency discriminates against words that are common across the corpus \cite{jurafsky2009speech,jones1972statistical}, which are assumed to be less representative of the document’s content, analogous to a set of domain-specific frequency-dependent stop words. The weighting $w_{t,d}$ obtained for each word of a document is used to identify the collection of key nouns to be retained. In particular, a tf-idf weighting threshold $\sigma_{tf-idf}$  is specified by the user (usually in terms of a user-defined percentile), and nouns with tf-idf weighting higher than the threshold are added to the collection of key nouns. 
 
\subsection{Relationship extraction} 
\label{subsection:relationshipextraction}
The input text is first tokenised into sentences using the NLTK package. After sentence tokenisation, open information extraction techniques are used to extract relationships from individual sentences \cite{jurafsky2009speech,mausam2016open}. 

\subsubsection{Open Information Extraction (IE)}
In this study, the OpenIE toolbox is used for relationship extraction \cite{christensen2011analysis,pal2016demonyms,saha2017bootstrapping,saha2018open}. The toolbox assigns a confidence value to each extracted relation, and in the proposed approach, a confidence threshold $\sigma_{{relationship}}$ between $0$ and $1$ is manually defined to preliminarily filter the extracted relationships. 

For each sentence, the toolbox generates a set of possible relations, where each relation is a list of textual phrases $r = (p_1,p_2,p_3,p_4,...,p_{N_r})$. In an extracted relation $r$, $p_2$ is the relation phrase, $p_1$ and $p_3$ are akin to subject and object phrases, and $p_4$ to $p_{N_r}$ are secondary argument phrases, sometimes appearing in long sentences, where $N_r$ is the number of phrases in an extracted relationship~\cite{mausam2016open}. The OpenIE toolbox is implemented as a combination of four methods \cite{christensen2011analysis,pal2016demonyms,saha2017bootstrapping,saha2018open}, which are briefly explained:
 
\textit{Semantic Role Labeling (SRL) based IE} SRL consists of detecting semantic arguments and their roles associated with a verb in a sentence. For example, given the sentence `I ordered a cake', SRL identifies `I' and `cake' as arguments for the verb `order', where `I' is the agent and `cake' is the patient. This component of the OpenIE tool is based on the SRL system by \cite{punyakanok2008importance} and \cite{johansson2008effect}.
 
\textit{Relational noun based IE} Relational noun based IE identifies relations that are mediated by nouns. For example, given the phrase `Rowing Club President James', it extracts the relation (James, be President of, Rowing Club). This is implemented by encoding a predefined set of relational nouns and nominal patterns in the OpenIE tool \cite{pal2016demonyms}. 
 
\textit{Numerical IE} Numerical IE identifies the numerical relations in a sentence. For example, given the sentence `The company has 100,000 employees', it extracts the relation (The company; has number of employees; 100,000). This is implemented by a combination of machine-learned patterns and customisations specific to numerical relations in the OpenIE tool \cite{saha2017bootstrapping}. 

\textit{Coordination analyser} This method is used to split conjunctive sentences. It first uses a dependency parser to extract candidate conjuncts, and then score their coherence based on the Berkeley Language Model \cite{pauls2011faster}. Additional linguistic constraints are also imposed to improve selection results \cite{saha2018open}.


\subsection{Key phrases and key relationships selection}
\label{subsection:keyselection}
To obtain the key phrases and key relationships useful for generating SysML model elements, a three-step selection mechanism is used. The first step is to select and refine candidate key phrases using the key nouns obtained in Section~\ref{subsection:nounextraction}. The second step is based on scoring the candidate phrases based on a combination of metrics to obtain key phrases~\cite{jurafsky2009speech,fellbaum98wordnet,WordNet}. The third step is to select key relationships based on the selected key phrases. 
 
\subsubsection{Selecting candidate key phrases}
\label{section:selectcandidatekeyphrase}
For simplicity, only subject and object phrases ($p_1$ and $p_3$) of the extracted relationships are used for identifying the key phrase candidates. The phrases are first preprocessed using procedures described in Section~\ref{section:preprocessing}. This results in tokenised phrases where $p = (t_1,t_2,...,t_{N_p})$, and each phrase comprises only nouns in root forms; $N_p$ refers to the number of nouns in a phrase. Then, an intersection is taken between each processed phrase and the list of key nouns. Nouns in these phrases that do not map to any key nouns are removed. The user is further able to determine the specificity of the key phrases by setting a parameter $L_{phrase}$ for the maximum number of nouns in one key phrase, where $length(p) \leq L_{phrase}$. If the number of nouns in a candidate key phrase exceeds $L_{phrase}$, the nouns are ranked according to tf-idf values and the top $L_{phrase}$ key nouns are kept. This results in a collection of candidate key phrases for further processing.

\subsubsection{Key phrase selection based on tf-idf, WordNet depth, and phrase frequency}\label{sec:WordnetTfIdf}
The candidate key phrases are further selected by using a metric based on the frequency of the phrase, average tf-idf weighting, and average WordNet score~\cite{fellbaum98wordnet,WordNet}. 

The count of a phrase, $count_{p,d}$, refers to the total count of a phrase $p$ in the `bag of phrases' of a document $d$, normalised with respect to the most frequent phrase in the document. This is used to account for the importance of each phrase. Only the candidate key phrases that are outputs of Section\ref{section:selectcandidatekeyphrase} are considered.

The WordNet depth $h'$ of a word $t$ refers to its semantic depth assigned by WordNet based on its synsets, and is used to account for semantic meaning when evaluating a phrase~\cite{fellbaum98wordnet,WordNet}. A general term such as `entity' would be assigned a small depth value, whereas a more specific term such as `pancreas' would be assigned a high depth value. The WordNet is a large lexical database of English, where the meanings of each word are represented as a cognitive synonym set called synsets~\cite{fellbaum98wordnet,WordNet}. The LESK algorithm is used to identify the most relevant synset for words in the input document based on the document as context \cite{lesk1986automatic}. For simplicity, it is assumed that each word in one document has only one synset. The WordNet depth of each word is normalised to range $0 \leq h' \leq 1$ by dividing by the largest depth value in a document, and the one complement of the normalised depth $h=1-h'$ is used as the WordNet score.

As both tf-idf weighting $w$ (already introduced in Section~\ref{section:evaluationoftfidf}) and the WordNet score $h$ are assigned to individual words $t$, an average value is taken for each candidate key phrase. Hence, for each candidate key phrase $p$, containing $N_p$ number of nouns, from a document $d$, a score $\lambda$ is computed by using the following formula:

\begin{equation}
\lambda_{p,d} = \sum\limits_{i}^{N_p}w_{t_i,d}/{N_p} + \sum\limits_{i}^{N_p}h_{t_i,d}/{N_p}  + \textup{count}_{p,d}    
\label{eq:entityscore}
\end{equation}

The first term computes the average tf-idf weighting, the second term computes the average WordNet depth, and the last term refers to the normalised count of the phrase. A score threshold $\sigma_{p}$ can then be specified by the user, where candidates with scores higher than the threshold are selected as key phrases. 
\textcolor{black}{It is worth mentioning that when documents contain many broken sentences (as in the document 7 of Table 2), many key phrases might not be found. Since this is an important step in the procedure, in order to support the engineers using the proposed tool, it is recommended to setup an alarm to warn the user when the tool fails to extract relations from more than 50\% of the sentence tokens in one document.} 

\subsubsection{Key relationship selection}
Relationships are selected as key relationships only when both their subject and object phrases are key phrases. This is to prevent open-ended relationships and `floating' blocks that do not connect to any other block in SysML diagrams.

\subsection{Mapping and augmentation}
\label{subsection:mappingandaugmentation}
The selected key phrases and key relationships are \textit{textual}. For plotting SysML diagrams, they need to be first mapped to SysML model elements. The authors focus on the Block Definition Diagrams (BDD), Internal Block Diagrams (IBD), and Requirement Diagrams (REQ) in this study. The key phrases are used to generate blocks $B$, which form a fundamental unit of SysML~\cite{friedenthal14practical}. The textual relationships are first classified into different categories according to the SysML diagram type, and then used to generate the different types of relationships $R$ between blocks. The textual relationships are also used to generate requirements that the blocks satisfy. 

To make diagrams more complete in both structure and semantics, the textual phrases and relationships are also augmented according to the required SysML diagram type. The augmentations are used to generate additional blocks and relationships for the diagrams. 

\subsubsection{Block Definition Diagrams}
\label{section:blockdefinitiondiagram}
Block Definition Diagrams are the most common diagrams used in SysML and are used to define the types of elements of other diagrams \cite{friedenthal14practical,omg19omg}. Blocks in BDD can have many features such as parts, references, and operations, whereas relations in BDD can belong to categories such as associations, generalisations, and dependencies \cite{friedenthal14practical,omg19omg}. For simplicity, the authors choose to focus on the composite association, generalisation, and reference association, and on identifying the operations features of blocks in BDD. The composite association relationship represents a structural composition where one block is part of another block \cite{friedenthal14practical,omg19omg}. On the other hand, generalisation relationship means an inheritance relationship, where one block is the generalisation of the other, more specialised, block \cite{friedenthal14practical,omg19omg}. Additionally, reference association conveys a connection between blocks where one block can access another \cite{friedenthal14practical,omg19omg}. 

\textit{Relationship mapping for BDD}

The steps used to map textual relationships to inter-block relationships in BDD are:
\begin{enumerate}
\item Identification of operation: for every subject phrase $p_1$ in the textual relationships, the relation phrase $p_2$ between the subject and its object phrase is considered as the operation of the block $B_{p1}$ defined by $p_1$. 
\item Classification based on relation phrase: the textual relationships are first classified according to the meaning of the relation phrase $p_2$ in the relationship. A pre-selected list of WordNet synsets that define composite associations are used, for example `include.v.01'. If the relation phrase’s synset belongs to the predefined list, then the relation is mapped to a composite association with the corresponding hierarchy.
\item Classification based on overlap: the rest of the textual relationships are then classified by the string representations of the subject and object phrases $(p_1,p_3)$. If the string representation of one phrase is contained in the other phrase in the relation (e.g. the phrase ‘prediction model’ includes the single word phrase ‘prediction’ in string representation), then the relationship is mapped to a generalisation relationship, with the block defined by the shorter phrase being the generalisation (the other being specialisation).
\item Classification based on score: the remaining textual relations are then classified according to the score of the subject and the object phrase according to Equation~\ref{eq:entityscore}. If the difference in score is above a user-defined threshold $\sigma_{rel-difference}$, the relation is mapped to a composite relationship where blocks defined by the lower-scored phrase is considered a part of the block defined by the higher-scored phrase.
\item Classification of remaining relations: the remaining textual relations which are not yet classified are then mapped to reference associations.

\end{enumerate}

\textit{Augmentation for BDD}

The steps used to augment the list of textual phrases and relationships for plotting BDDs are:
\begin{enumerate}
\item Identification of top-level phrases: The key phrases whose corresponding blocks do not form sub-blocks of another are first compiled to a set $P$, where a sub-block is defined as the block at the part end of a composite relationship, or at the specialised end of a generalisation relationship. These phrases are used for the next abstraction step.
\item Abstraction: Algorithm~\ref{algo:entityabstraction} is used to iteratively abstract higher-level phrases from the top-level phrases based on a per-word score $\gamma$ that is a combination of the tf-idf weighting $w$ and WordNet score $h$ of a word $t$ in a document $d$. 

\begin{algorithm}[h]
Identify set of top-level phrases $P$ \\
\SetAlgoLined
 \For {\textup{phrase} p \textup{\textbf{in}} P} {
  \If{\textup{length of phrase} $N_p >$ \textup{1}}{
    Initialise score set $\Gamma$ \\
    \For {\textup{noun} t \textup{\textbf{in}} p}{
        $\gamma_{t,d} = w_{t,d} + h_{t,d}$
        
        $\Gamma_{p,d} = \Gamma_{p,d} \cup \{\gamma_{t,d}\}  $
        }
    sort($\Gamma_{p,d}$)
    
    $p_{abstract} = p\setminus t$, where $\gamma_{t,d}$ is smallest in $\Gamma_{p,d}$ 

	$P = P \cup \{p_{abstract}\}$
}}
\caption{Phrase Abstraction}
\label{algo:entityabstraction}
\end{algorithm}
Additionally, generalisation relationships are assigned to the blocks defined by the top-level phrases and their abstractions. 

\item Relationship augmentation: After the abstraction step, all of the resultant sets of top-level phrases are now unigrams. The WordNet is then used to identify hypernym/hyponym (corresponding to generalisation) and holonym/meronym  (corresponding to composite association) relationships between these phrases and corresponding relationships are assigned between their blocks~\cite{fellbaum98wordnet,WordNet}.

\item Phrase augmentation: As the final step, the lowest common hypernyms between the phrases that remain at the top level are found using WordNet ~\cite{fellbaum98wordnet,WordNet}. The blocks defined by these remaining top-level phrases are assigned a generalisation relationship with blocks defined by their hypernyms. 
\end{enumerate}

\subsubsection{Internal Block Diagrams} 
Internal Block Diagrams are used to specify the internal structures of blocks, and display the parts and references of a specific block as well as the connections between its parts and references \cite{friedenthal14practical,omg19omg}. In this study, the parts of a block are defined as the block's composites, and the references of a block are defined as other blocks connected to it through the reference association. A connection means that two blocks are able to access each other in an operational system \cite{friedenthal14practical,omg19omg}. To leverage the connection between IBD diagrams and BDD diagrams, the augmented list of textual relationships and phrases from the BDD are used as inputs to the IBD algorithm.

\textit{Relationship mapping for IBD}
 
The steps to map textual relationships to inter-block relationships for IBD are: 
\begin{enumerate}
\item Parent block generation: The user can first choose to specify a parent block to draw the IBD for. Then, all blocks that are sub-blocks or sub-sub-blocks of the user-specified block are selected. If no parent block is manually selected, IBDs will be drawn for all blocks.

\item Relationship selection: The textual relationships whose subject or object phrases correspond to a sub-block of the user-specified block are selected, except in cases where the phrase is in the lower hierarchy in the textual relationship.

\item Connection classification: For each of the selected textual relationship, if the relation phrase is not empty, then blocks defined by its subject and object phrases are assigned a port connection between each other.

\end{enumerate}

\textit{Augmentation for IBD}

The steps used to augment the list of textual phrases and relationships for plotting IBDs are:

\begin{enumerate}
\item Phrase augmentation: Pairs of key phrases that share a non-empty intersection (i.e. include one or more identical nouns) are identified and are added to the list of phrases for IBD, except when the intersection corresponds to the user-specified block.

\item Relationship augmentation: After the phrase augmentation step, a generalisation relationship is assigned to the block defined by the intersection and the intersection's parent blocks, with the intersection block at the generalised end. The new relationships are added to the list of IBD relationships.
\end{enumerate}

Importantly, the augmentation mechanism used in IBD is different to the abstraction mechanism in BDD. This is done as a redundancy measure to ensure the completeness of the plotted diagrams. For example, a phrase ‘prediction mode’ may be only abstracted to ‘prediction’ in the BDD step. However, in the IBD step, it could be additionally abstracted to ‘model’ if any of the other selected key phrases also include the word ‘model’.
 
\subsubsection{Requirement Diagram}
Requirement Diagrams are a unique feature of SysML diagrams that represent the requirements of the system and its components \cite{friedenthal14practical,omg19omg}. Relationships relevant to requirements in SysML include containment, trace, derive, refine, satisfy, and verify \cite{friedenthal14practical,omg19omg}. For simplicity, the authors focus on the satisfy relationships and the trace relationships between the requirements, which represents a weak dependency where a change at one end may result in the need to modify the requirement at the other end \cite{friedenthal14practical,omg19omg}. 

\textit{Mapping for REQ}

The following steps are used to map textual relationships to requirements and requirement relationships for REQ diagrams.

\begin{enumerate}
    \item Requirement identification: For each of the key relationships, if the relation phrase is not empty, then its corresponding raw relationship output from OpenIE is considered a requirement. 
    \item The requirements are assumed to be satisfied by their subject phrase blocks, and their requirement blocks are named after the subject phrases. There can be multiple requirements with the same subject phrase, and these are grouped together in the same requirement block.
    \item Relationship identification: A trace relationship is assigned to requirements whose corresponding relationships share the same subject phrases or object phrases, or have subject phrases equal to object phrases of other relationships, and vice versa. 
\end{enumerate}

\textit{Relationship augmentation for REQ}

The WordNet is used to identify any hypernym, hyponym, entailment, and causes relationship between the relation phrases of relationships that have been identified as a requirement ~\cite{fellbaum98wordnet,WordNet}.  If a relationship is identified, the corresponding pair of requirements are considered to have a trace relationship
 
\subsection{Diagram Generation}
\label{subsection:diagramgeneration}
Prior to plotting the diagram, the user can choose to select a parent block for plotting the desired diagram. For IBD, this would be the parent block used to select sub-blocks and sub-sub-blocks. For BDD, all blocks at a lower hierarchy than the parent block (e.g. the parent block's sub-blocks and their sub-blocks, etc.) will be iteratively selected, and the resultant selected blocks and their relationships will be plotted instead of the full diagram comprising all generated blocks. For REQ, all requirements that contain the phrase corresponding to the selected block, and the relationships between these requirements, are extracted for plotting. 

An open-source diagram generation tool (PlantUML~\cite{plantuml,plantumlguide}) is used to plot the generated blocks and their relationships. The augmented blocks and relationships are plotted with dotted lines to indicate that these do not directly map to phrases and relationships in the input text documents. The diagram generation tool uses GraphViz as its graphical engine \cite{ellson2002graphviz}. 

\section{Steps for the implementation of the proposed approach}
\label{section:implementation}

The following steps are performed for the generation of SysML diagrams using the proposed approach:

\begin{enumerate}
\item Select the corpus of textual materials;
\item Select the document to draw the SysML diagram for;
\item Set threshold values for five hyperparameters
    \begin{enumerate}
    \item $0<\sigma_{tf-idf}<1$ for key noun selection. This sets the tf-idf threshold for a noun to be considered a key noun.
    \item $0<\sigma_{{relationship}}<1$ for relationship extraction. This sets the confidence threshold for a relation extracted by OpenIE and can be used to remove an excessive number of duplicated relationships.
    \item $0<\sigma_{{p}}<3$ for key phrase selection. This sets the minimum score for a phrase to be considered a key phrase via the proposed formula in Equation~\ref{eq:entityscore}.
    \item $0<\sigma_{{rel-difference}}<3$ for composite relationship classification. This sets the minimum score difference for connected phrases to be mapped to composite relationships.
    \item $L_{{phrase}}>0$ for length of phrases. This sets the maximum number of words in a phrase;
    \end{enumerate}
\item Select the type of SysML diagram to generate (optional);
\item Select the phrase to define the parent block in the SysML diagram (optional);
\item Run the end-to-end six-steps approach as described in Section~\ref{section:methods} to automatically generate SysML diagrams, which will generate the following outputs in sequence:
    \begin{enumerate}
    \item List of key nouns;
    \item List of extracted relationships;
    \item List of key phrases and key relationships;
    \item Mapping of key textual relationships to SysML relationships;
    \item Required diagrams.
    \end{enumerate}
\item Evaluate the quality of the results through the following steps:
    \begin{enumerate}
    \item Evaluation of key phrase selection;
    \item Evaluation of key relation mapping;
    \item Evaluation of diagram generation.
    \end{enumerate}
\end{enumerate}

\subsection{Quantification of the success of phrase extraction}
To evaluate the success in key phrase extraction, manually selected lists of key phrases based on OpenIE extractions are created for each of the selected documents and used as the ground truth. The automatically extracted lists of phrases are then evaluated by precision and recall. Precision is defined as the percentage of extracted key phrases that match a phrase in the ground truth. Recall refers to the percentage of ground truth phrases that match a phrase in the extracted list. As the key relationships are extracted together with the key phrases, the key relationships are not evaluated separately because it is strongly coupled with the key phrases. 

\subsection{Quantification of the success of relationship mapping}

To evaluate the effectiveness of the approach in mapping textual relationships to SysML relationships, the ground truth for the type of relationship between blocks defined by the subjects and objects is defined manually. This is used as a benchmark to evaluate the results from the relationship mapping algorithm. The algorithm is evaluated for its accuracy in determining the type of SysML relationship (e.g. composite association, generalisation or reference association) and the relative hierarchy between the blocks defined by the subject and the object phrase in the textual relationship (e.g. whether block defined by the subject phrase is a composite of that defined by the object, or vice versa).

\subsection{Hyperparameter selection}
The following values for the hyperparameters are suggested, and applied to the six case studies investigated:
\begin{enumerate}
    \item $\sigma_{tf-idf}=0$. This means that all the nouns in the input text are treated as key nouns. This is done to illustrate the effect of selection in the subsequent steps. However, it is envisioned that the parameter can be adjusted according to the user’s needs;
    \item $\sigma_{{relationship}}=0.5$. This is to filter out relationships with low confidence of being valid relationships.
    \item $\sigma_{{p}}=0.6$. This implies most phrases are selected as key phrases, to ensure the completeness of results. 
    \item $\sigma_{{rel-difference}}=0.5$. This ensures that only connected phrases with a sufficient score difference are mapped to composite relationships.
    \item $L_{{phrase}} = 3$. This sets the maximum length of a phrase to be three nouns. 
\end{enumerate}

\subsection{User Interface}

\begin{figure}[h]
    \centering

    \includegraphics[width=0.9\columnwidth]{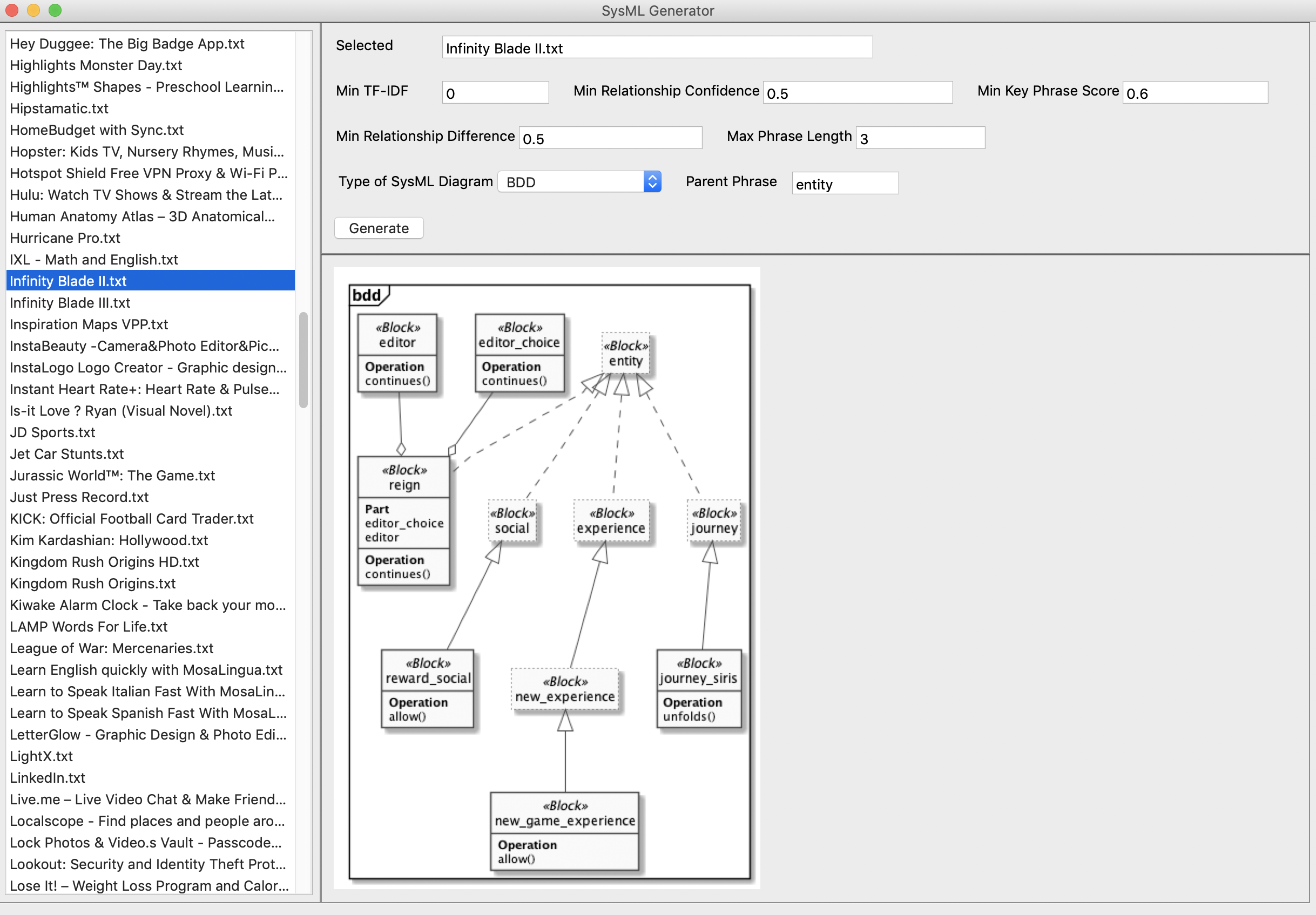}

\caption{An example of the user interface for the proposed tool}
\label{figure:ui}

\end{figure}

\textcolor{black}{An illustrative example of the user interface for the proposed tool is shown in Figure~\ref{figure:ui}. The main panel features the list of uploaded documents on the left. On the top, the users are able to vary the hyperparameters according to need, and generate the desired SysML diagram at the bottom, below the hyperparameters. }

\section{Experiments}
\label{section:exp}

Six case studies are investigated to evaluate the effectiveness of the proposed approach in automatically generating SysML diagrams from texts. The details of the datasets used, the experimental procedure, the evaluation procedure, and the choice of hyperparameters are discussed in this section. The results from the extraction of key phrases and classification of relationships are validated against manually constructed benchmarks. Additionally, the generated diagrams are also compared against manually designed ground truth diagrams.

It is worth emphasising here that the aim of the proposed approach is not to replace systems engineers but rather to aid them in gaining an overview of the system. Therefore, the goal is not a perfect extraction of phrases or relationships. Rather, it is to provide an overview of the selected document and system through the profiles in SysML diagrams.

\subsection{Datasets}

The applicability of the proposed approach is tested on six case studies to cover different dataset sizes, domains, and writing styles:

1) Patent descriptions from the European Patent Office (EPO)~\cite{epopatent}. Patent files in English are downloaded from the EPO public database in txt format. Each patent includes sections such as patent name, abstract, and patent description. The patent files are first parsed to remove constructs such as HTML tags and figure references. Then, only the patent description section is used as textual input to the algorithm. Each description is treated as a separate document, and 345 separate patent description documents are obtained. The patents cover areas such as data transmission devices, devices for drug admission, and bioinformatics. 

2) Windows manual. A digital book named Windows 10 Field Guide in pdf format is used~\cite{windows10fieldguide}. The book is split into 28 sections according to the chapter, and converted to individual txt files with images removed. Each file is treated as a separate document. The book is a manual for the Microsoft operating system Windows 10 and describes its different functions and applications, for example Skype and Maps.

3) UK government report. A series of UK government report on the Future of Manufacturing is downloaded as pdf files ~\cite{ukgovreport}. A total of 37 reports are downloaded, and each of them is further split into different chapters, with cover pages, acknowledgements, content pages and references removed. This results in a total of 284 chapters. Each chapter is converted to txt format and treated as a separate document. The contents cover areas such as technology trends in manufacturing and investment in innovation. 

4) App descriptions. App descriptions from the Apple App Store are downloaded from an online public dataset~\cite{appstore}. A total of 346 app descriptions in txt format with a file size larger than 4KB are selected from a dataset of 4075 app descriptions. This is done to ensure that each document has sufficient length. Each description is treated as a separate document. The app descriptions included apps of different genres and included contents such as user reviews and feature descriptions.

5) Research publications from IEEE. 285 papers are downloaded using IEEE Xplore ~\cite{ieeexplore}. The search criteria were to have the words ‘design’ and ‘manipulator’ in the title of the paper and the range was from 2015 to 2021. The papers were downloaded as pdfs and converted to txt files. Each paper is treated as a separate document. The papers covered areas in the design of robotic manipulators, for example mechanics and control. 

6) Description of countries on Wikipedia. 193 English Wikipedia pages about countries are parsed from the parent Wikipedia page \textit{Member states of the United Nations}~\cite{wikipedia}. Each page is saved as a separate txt file and all formatting and images were removed. Each page is also treated as a separate document. The descriptions include each country's economy, geography, government, etc. 

Details of each dataset are summarised in the table below. Two documents are selected from each of the datasets for demonstration\footnote{The demonstration documents are available at https://github.com/ShaohongZ/NLP-for-Systems-Engineering}. The documents are selected based on their word counts being close to the average word count of the dataset. Additionally, one block is selected for each document as the parent block, and the phrases that correspond to the parent blocks are listed in Table~\ref{table:datasets}. The use of these phrases is explained in Section~\ref{subsection:diagramgeneration}.

\begin{table}[ht!]
\centering
\begin{tabular}{M{6cm}|M{1cm} M{3.5cm} M{2,5cm}} 
& Doc. no. & Word count & Selected phrase \\ \hline
\textbf{EPO patent}  & & 8370 (average) & \\ \hline
A state estimator & 1 & 8418 & Prediction \\ \hline
A body fluid leakage detection\newline aqueous composition & 2 & 8343 & Starch \\ \hline

\textbf{Windows manual}  & & 3027 (average) & \\ \hline
Devices & 3 & 3061 & Display \\ \hline
Maps & 4 & 2859 & Map \\ \hline 

\textbf{UK government report}  & & 1982 (average) & \\ \hline
Knowledge spillover - knowledge sources manufacturing Part5 & 5 & 1993 & Intangible asset \\ \hline
De-industrialisation and balance-of payments Part3 & 6 & 1958 & Manufacturing \\ \hline

\textbf{App description}  & & 559 (average) & \\ \hline
Infinity Blade II & 7 & 559 & Entity \\ \hline
Ebates: Cash Back, Coupons \& Rebate Shopping App & 8 & 559 & Entity \\ \hline

\textbf{Research paper}  & & 3414 (average) & \\ \hline
Design of a weight-compensated and coupled tendon-driven articulated long-reach manipulator & 9 & 3430 & Actuator \\ \hline
Multi criteria design of a spherical 3-DoF parallel manipulator for optimal dynamic performance & 10 & 3413 & Optimisation \\ \hline

\textbf{Wikipages}  & & 9830 (average) & \\ \hline
Bhutan & 11 & 9867 & Industry \\ \hline
India & 12 & 9811 & Religion \\ \hline
\end{tabular}
\caption{Details of datasets, selected documents, and selected phrases}
\label{table:datasets}
\end{table}







\section{Results}
\label{section:results}

\subsection{Key phrase extraction}

The results from the key phrase extraction are detailed in Table~\ref{table:keyentity}. It can be seen that the number of phrases extracted from a document is around the same order of magnitude as the number of unique nouns in the document. The possible reason why the number of phrases tends to be less than the number of nouns, except for one case, is that many of the phrases are in descriptive sentences or sentences with pronouns. For example, in a descriptive sentence such as `The device is expensive', the candidate phrase `device' will not be considered as a key phrase as no relationship can be extracted from the sentence. Additionally, in sentences that involve pronouns such as `You can turn on the computer' and `It is used to predict motion', the candidate phrase `computer' or `motion' will not be selected because detecting pronoun references is out of the scope of this paper. Requiring all key phrases to be related to at least one other phrase is useful because it avoids the case of `floating blocks' where a given block is not connected to any other block in the SysML diagram. Such floating blocks are not conducive to helping the user understand the dependencies within the system or the hierarchies between the different blocks within the system, and may confuse the user. 

\begin{table}[ht!]
\centering
\begin{tabular}{M{2cm}|M{2cm} M{2cm} M{2cm} M{2cm} M{2cm}} 
\makecell{Document\\No.} & \makecell{Word\\Count} &\makecell{Unique\\nouns}& \makecell{No. of\\key phrases}& Precision \% & Recall \% \\ \hline
1  & 8418 & 345  & 401 & 80.8 & 76.8 \\ \hline
2  & 8343 & 430  & 351 & 91.5 & 77.3 \\ \hline
3  & 3061 & 214  & 141 & 70.2 & 57.9 \\ \hline
4  & 2859 & 193  & 119 & 66.4 & 49.7 \\ \hline
5  & 1993 & 214  & 157 & 55.4 & 81.3 \\ \hline
6  & 1958 & 223  & 140 & 70.7 & 71.2 \\ \hline
7  & 559  & 122  & 12  & 66.7 & 34.8 \\ \hline
8  & 559  & 105  & 33  & 84.8 & 63.6 \\ \hline
9  & 3430 & 216  & 131 & 61.8 & 54.0 \\ \hline
10 & 3413 & 367  & 211 & 81.0 & 82.6 \\ \hline
11 & 9867 & 1198 & 992 & 77.3 & 72.9 \\ \hline
12 & 9811 & 1224 & 854 & 58.7 & 69.3 \\ \hline
\end{tabular}
\caption{Results from key phrase extraction}
\label{table:keyentity}
\end{table}

As can be seen from Table~\ref{table:keyentity}, the precision and recall rates indicate that the algorithm has successfully extracted key phrases from the document. The algorithm is especially useful with documents that tend to be more carefully written, as can be seen from the higher success rates in extracting phrases from patent descriptions. The failure case of key phrases that are not identified includes those in descriptive sentences, those connected to pronouns, and incomplete sentences. Incomplete sentences are probably the main reason for the low recall in extracting key phrases from Document No.7 because the descriptions contained many broken sentences such as `The best game ever'. Additionally, the algorithm is able to achieve relatively high precision, across most of the document types, validating the proposed algorithm for selecting key phrases. The precision and recall rates can be further improved by adjusting the hyperparameters. The success in extracting key phrases contributes to helping the user gain an initial understanding of the system.

\textcolor{black}{As no other approach automatically generates SysML diagrams directly from text, the comparison with the literature can only be made by considering the extraction of key phrases for similar purposes. It was observed, that the precision and recall results are comparable with or outperform most prior works. Specifically in~\cite{sawyer2005shallow}, which aimed to extract domain terms as UML classes, a precision of 40\% and recall of 50\% were reported. In the ontology domain, the CRCTOL~\cite{jiang2010crctol} reports more than 90\% in precision and less than 5\% in recall (in different ablations). The proposed method also achieves comparable results to~\cite{bakar2016extracting} (precision of 62\% and recall of 82.2\%), \cite{sreekumar2018extracting} (precision between 40\% and 73\%, and recall between 57\% and 93\%), and \cite{hariri2013supporting} (precision between 82.5\% and 85.0\%, and recall between 68.0\% and 53.2\%).}

\textcolor{black}{On the other hand, the proposed approach under-performs compared to \cite{robeer2016automated} (precision and recall rates around 90\%) and~\cite{afreen2011generating} (precision of 83.82\% and 91.01\%). This is because, \cite{robeer2016automated} assumes a fixed format for the user stories, thus making the extraction process much simpler. Similarly, \cite{afreen2011generating} assumes a structure in the input text, arguably simplifying the extraction. }

\begin{table}[htp!]
\centering
\begin{tabular}{M{2cm}|M{3cm} M{3cm} M{3cm}} 
\makecell{Document\\No.} & \makecell{No. of\\sentences} &\makecell{No. of\\extracted relations}& \makecell{No. of\\key relations} \\ \hline
1  & 409 & 670  & 425 \\ \hline
2  & 463 & 681  & 309 \\ \hline
3  & 186 & 257  & 102 \\ \hline
4  & 174 & 289  & 107 \\ \hline
5  & 92  & 286  & 137 \\ \hline
6  & 98  & 158  & 100 \\ \hline
7  & 20  & 21   & 7   \\ \hline
8  & 34  & 43   & 25  \\ \hline
9  & 301 & 251  & 115 \\ \hline
10 & 217 & 207  & 150 \\ \hline
11 & 508 & 1179 & 799 \\ \hline
12 & 443 & 1106 & 689 \\ \hline
\end{tabular}
\caption{Results from key relationship extraction}
\label{table:keyrelationshipextraction}
\end{table}

\subsection{Key relationship extraction}
The results from key relationship extraction are detailed in Table~\ref{table:keyrelationshipextraction}. As can be seen from the table, the OpenIE tool extracts multiple semantic relations from the same sentence even after applying a relation likelihood threshold of 0.5 on the OpenIE tool. This is likely attributed to the combination of different methods used in OpenIE \cite{christensen2011analysis,pal2016demonyms,saha2017bootstrapping,saha2018open}. This is helpful in capturing the complete semantic meaning of each sentence. It can also be seen that the key relation selection algorithm is effective in removing from 27\% to 65\% of the extracted relations. This can also be manually adjusted by tuning the parameters $sigma_{tf-idf}$, $\sigma_{p}$, and $\sigma_{{relationship}}$. The use of multiple parameters is aimed at giving the user more control over the completeness of the generated SysML diagrams.

\subsection{Relationship mapping}

The results from the mapping of key relationships for BDD diagram are detailed in Table~\ref{table:keyrelationshipclassification}. A majority of the extracted textual relationships are mapped to either composite associations or reference associations. This is expected because a textual relationship is only mapped to generalisation if one phrase contains another, which is assumed to be a stricter criterion. At a difference threshold of 0.2, the number of composite associations and the number of reference associations are around the same order of magnitude, with the number of composite associations generally larger. \textcolor{black}{At a difference threshold of 0.5, the number of composite and reference associations are around the same order of magnitude, with the number of reference associations generally larger. } 

\begin{table}[ht!]
\centering
\begin{tabular}{M{2.2cm}|M{2.2cm} M{2.2cm} M{2.2cm} M{2.2cm}} 
\makecell{Document\\No.} & Generalisation &\makecell{Composite\\association}& \makecell{Reference\\association} & \makecell{Classification\\accuracy \%} \\ \hline
1  & 26 & 168 & 226 & 75.8 \\ \hline
2  & 11 & 92 & 204 & 70.9 \\ \hline
3  & 3  & 31  & 68  & 64.7 \\ \hline
4  & 6  & 45  & 55  & 70.1 \\ \hline
5  & 4  & 39  & 93  & 84.7 \\ \hline
6  & 1  & 26  & 72  & 71.0 \\ \hline
7  & 0  & 2   & 5   & 57.1 \\ \hline
8  & 0  & 11  & 13  & 68.0 \\ \hline
9  & 7  & 34  & 73  & 62.6 \\ \hline
10 & 1  & 33  & 115  & 73.3 \\ \hline
11 & 12  & 209 & 577 & 73.0 \\ \hline
12 & 5  & 187 & 497 & 73.6 \\ \hline
\end{tabular}
\caption{Results from key relationship classification}
\label{table:keyrelationshipclassification}
\end{table}

Having a sufficient number of composite relationships and generalisation relationships is useful because they also serve as taxonomic relationships which provide structures to the SysML diagrams. This helps the algorithm to identify the hierarchy between blocks and facilitates the graphic layout algorithm. Having such a hierarchical structure is useful in helping the user understand the overall structure of the system. Additionally, the algorithm can also be seen to have achieved high accuracy in mapping the correct relationship types. The outlier in the app description dataset is potentially due to the small number of key relations for the document. This contributes to creating a more objective starting point in SysML diagrams for users. 

\textcolor{black}{Due to the differences in the proposed approach, comparisons are made with methodologies that aim to leverage NLP techniques for extracting/classifying requirements and relationships. For extracting relationships, the results are comparable to those reported in ~\cite{jiang2010crctol} (F-score of around 69\%) and ~\cite{sreekumar2018extracting} (F-score from 48.0\% to 81.1\%). Similar results were also obtained in~\cite{drymonas2010unsupervised}, reporting results in taxonomy construction with F-scores of between 66.7\% and 69.2\%. The proposed approach under-performs compared to~\cite{robeer2016automated}, which reports a precision between 67.2\% and 83.7\% in extracting relationships (F-score$\sim$80\%). This is because in \cite{robeer2016automated} is assumed a fixed input format, thus making the extraction process simpler. Similarly, \cite{nguyen2015rule} reports results with F-score of 80.3\%, by making assumptions on the structure of the input documents. Additionally, when considering requirement classification results, the results obtained are comparable to~\cite{casamayor2010identification} that uses 5\%-15\% of the dataset for training.  }

\subsection{Augmentation}

The results from augmentation for BDD diagrams are detailed in Table~\ref{table:augmentation}. As can be seen, the abstraction step adds a high number of augmented relationships and phrases. It helps provide a higher-level overview of the entire system and more structure to the SysML diagrams through taxonomic generalisation relationships. This also enables the user to quickly identify higher-level phrases within the document to plot other diagrams, for example package diagrams. Additionally, abstraction is based on the extracted phrases, ensuring that the abstracted phrases and relationships are relevant to the original corpus. This provides users with a more complete starting point to understand and design the complex system. 

\begin{table}[ht!]
\centering
\begin{tabular}{M{2cm}|M{2cm} M{2cm} M{2cm} M{2cm} M{2cm}} 
\makecell{Document\\No.} & \makecell{No. of\\abstraction\\relations} &\makecell{No. of\\abstracted\\phrases}&\makecell{No. of\\augmented\\top-level\\relationships}& \makecell{No. of\\lowest\\common\\hypernym\\relationships} & \makecell{No. of\\lowest\\common\\hypernyms} \\ \hline
1  & 246 & 125 & 1  & 276  & 96  \\ \hline
2  & 224 & 145 & 3  & 569  & 168 \\ \hline
3  & 98  & 61  & 1  & 159  & 58  \\ \hline
4  & 66  & 45  & 0  & 112  & 53  \\ \hline
5  & 104  & 62  & 1  & 235  & 85  \\ \hline
6  & 107  & 69  & 0  & 249  & 80  \\ \hline
7  & 10   & 10   & 0  & 14   & 9  \\ \hline
8  & 19  & 16  & 0  & 43   & 23  \\ \hline
9  & 83  & 51  & 2  & 121  & 51  \\ \hline
10 & 182 & 117  & 0  & 316  & 104  \\ \hline
11 & 736 & 462 & 26  & 1678  & 403 \\ \hline
12 & 595 & 376 & 37 & 1672 & 399 \\ \hline
\end{tabular}
\caption{Results from phrase and relationship augmentation}
\label{table:augmentation}
\end{table}

It can be seen that only a small number of relationships are identified between the top-level unigram phrases after the abstraction step, potentially due to the criteria used as only WordNet hypernyms/hyponyms and meronyms/holonyms relationship are identified~\cite{fellbaum98wordnet,WordNet}. It can also be seen that a large number of common hypernyms and hypernym relationships are augmented. The common hypernyms are able to provide another level of abstraction to the selected phrases based on semantics that are different from the abstraction based on lexical terms, helping the system engineer to gain a more complete high-level picture of the system and its different abstract categories in addition to contributing more structure to the SysML diagrams. 

\subsection{Requirement identification}
The results from requirement identification are detailed in Table~\ref{table:requirement}. The number of requirements identified is the same as the number of key relationships, which is expected because the requirements in this study are derived from relationships. This contributes to a more complete set of candidate requirements for the user to consider. Additionally, it can also be seen that a large number of relationships between requirements are identified and augmented through the algorithm. This leads to most of the requirements being connected using the augmented relationships, which is useful because even though relationships between phrases may be found through using sentences explicitly stated in the text, relationships between requirements, which are derived from sentences, can be difficult to identify. The proposed algorithm is able to suggest candidate relationships between requirements, allowing the user to build on the knowledge to classify relationships between requirements further. Additionally, the relationships also provide structure to the requirement diagrams by connecting the requirements and bringing related requirements to the same cluster.  

\begin{table}[ht!]
\centering
\begin{tabular}{M{2cm}|M{4cm} M{4cm}} 
\makecell{Document\\No.} & \makecell{No. of\\requirements} &\makecell{No. of relationships\\between requirements} \\ \hline
1  & 425 & 636 \\ \hline
2  & 309 & 450 \\ \hline
3  & 102 & 84 \\ \hline
4  & 107 & 52 \\ \hline
5  & 137 & 85 \\ \hline
6  & 100 & 56 \\ \hline
7  & 7   & 1 \\ \hline
8  & 25  & 2 \\ \hline
9  & 115 & 149 \\ \hline
10 & 150 & 161 \\ \hline
11 & 799 & 469 \\ \hline
12 & 689 & 575 \\ \hline
\end{tabular}
\caption{Results from requirement identification and augmentation}
\label{table:requirement}
\end{table}

\subsection{BDD diagram generation}
A few examples of BDD diagrams generated in the case studies are illustrated in Figure \ref{fig:BDD_diagrams} below. It can be seen that the augmentation steps were useful in building the structure of the BDD diagrams. For example, many blocks have corresponding phrases that share the same abstracted phrase, which defines the parent block that clusters these blocks together. This can also be tuned by the choice of specificity $L_{phrase}$. This is useful because these phrases are not explicitly related in the original document, yet they are connected via the parent blocks in the generated diagrams. For example, in Figure \ref{figure:bdd_display}, the block defined by the abstracted phrase `display' clusters the sub-blocks such as `display option' and `taskbar display' into one branch. In Figure \ref{figure:bdd_prediction}, the block defined by the abstracted phrase `prediction' clusters multiple sub-blocks into one branch. Additionally, blocks corresponding to top-level unigram phrases can also be connected to the same block defined by their common hypernyms such as `entity' in Figure \ref{figure:bdd_entity}, where `entity' is a high-level common hypernym for many synsets in the WordNet ~\cite{fellbaum98wordnet,WordNet}. 

\newpage
\begin{figure}[ht!]
    \centering
    \begin{subfigure}[b]{\columnwidth}
    \centering
    \includegraphics[width=0.95\columnwidth]{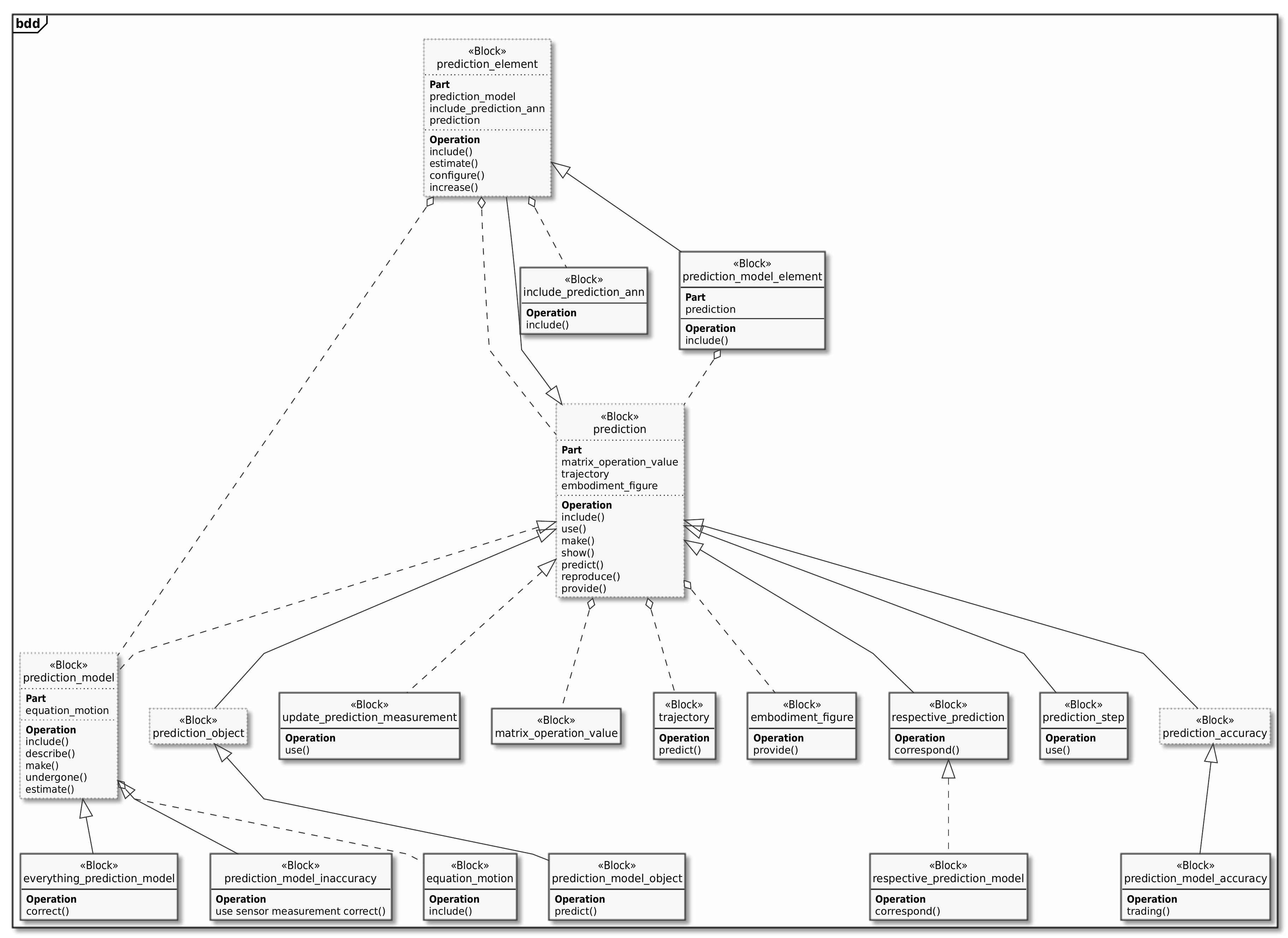}
    \caption{EPO patent dataset, selected phrase = `prediction'
    }
    \label{figure:bdd_prediction}
    \end{subfigure}

    \begin{subfigure}[b]{\columnwidth}
    \centering
    \includegraphics[width=0.95\columnwidth]{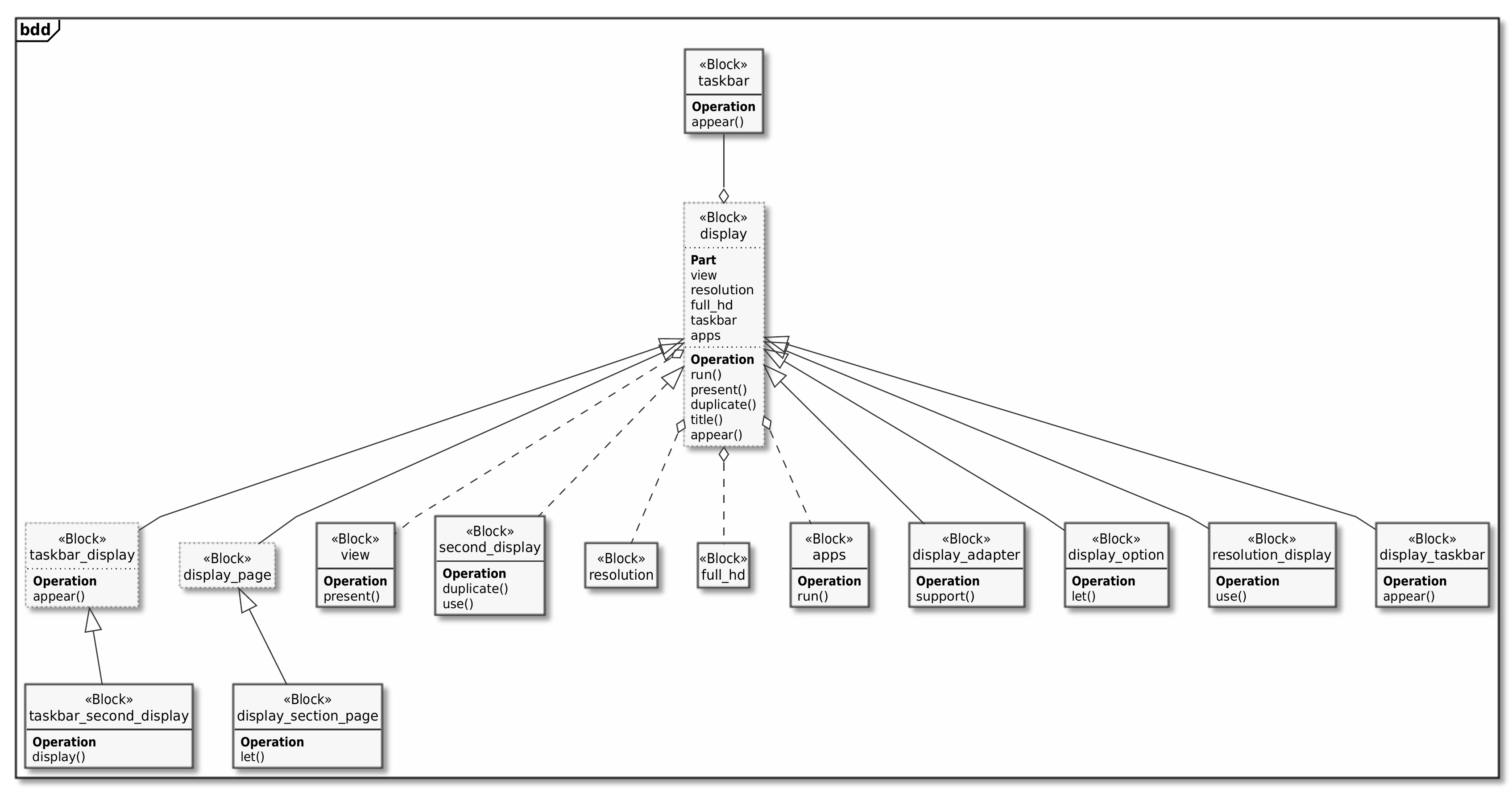}
    \caption{Windows manual dataset, selected phrase = `display'}
    \label{figure:bdd_display}
    \end{subfigure}
\end{figure}

\newpage
\begin{figure}[ht!]\ContinuedFloat
    \centering
    \begin{subfigure}[b]{0.6\columnwidth}
    \centering
    \includegraphics[width=0.6\columnwidth]{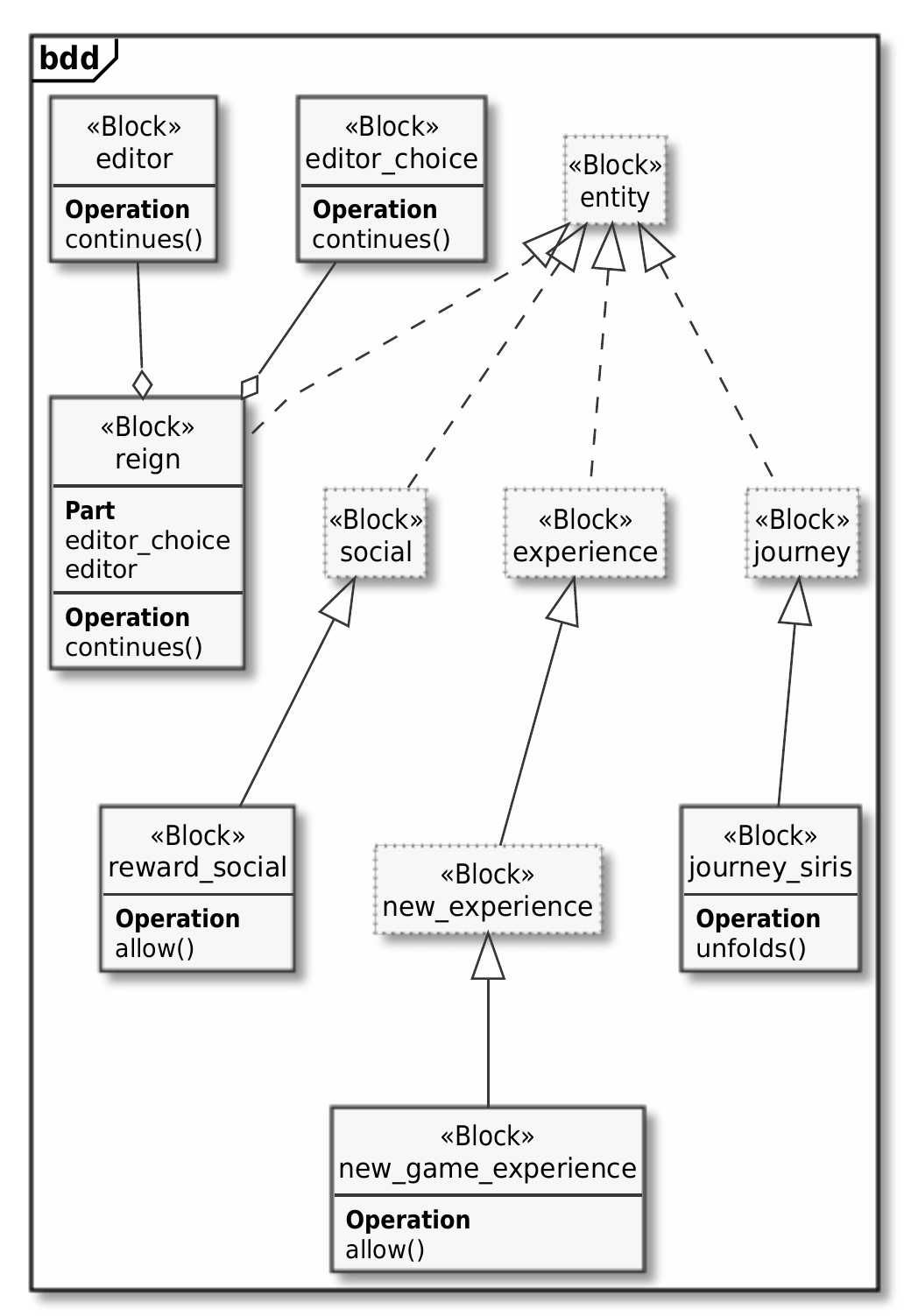}
    \caption{App description dataset, selected phrase = `entity'}
    \label{figure:bdd_entity}
    \end{subfigure}
    
    \begin{subfigure}[b]{0.25\columnwidth}
    \centering
    \includegraphics[width=\columnwidth]{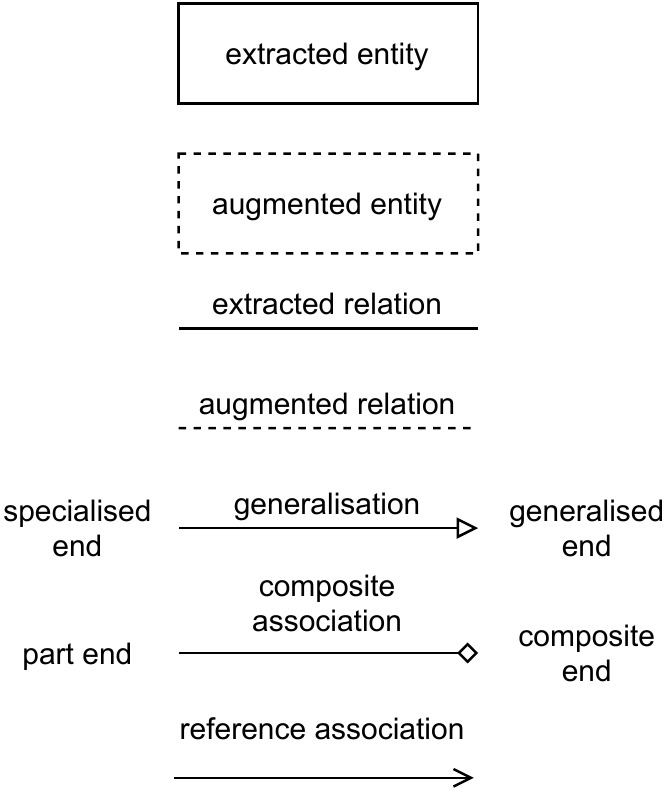}
    \end{subfigure}
    
\caption{Automatically generated BDD diagrams}
\label{fig:BDD_diagrams}
\end{figure}

These high-level phrases are augmented based on semantics, which ensures that they are relevant to the document, in addition to providing a hierarchical structure that aids user understanding. Without the augmented phrases, these extracted key phrases may become open-ended, which makes it harder for the user to understand the structure of the system from the SysML diagrams. Additionally, the two levels of abstraction capabilities, both lexical and semantic, are useful to help the systems engineer in gaining an overview of the higher-level abstract concepts in the system that are not explicitly stated in the text, and offer suggestions for candidate high-level phrases and their corresponding blocks to the systems engineers. 

It can also be seen that there is a mix of extracted relationships and augmented relationships in the diagram, which implies that both steps are useful for the construction of the SysML diagrams. By combining the taxonomy derived from both generalisation and composite association relationships and both extracted and augmented relationships, the algorithm ensures a strong structure in the generated diagrams. 

\subsection{IBD diagram generation}
\begin{figure}[p]
    \centering
    \begin{subfigure}[b]{\columnwidth}
    \centering
    \includegraphics[angle=270,width=5.8cm]{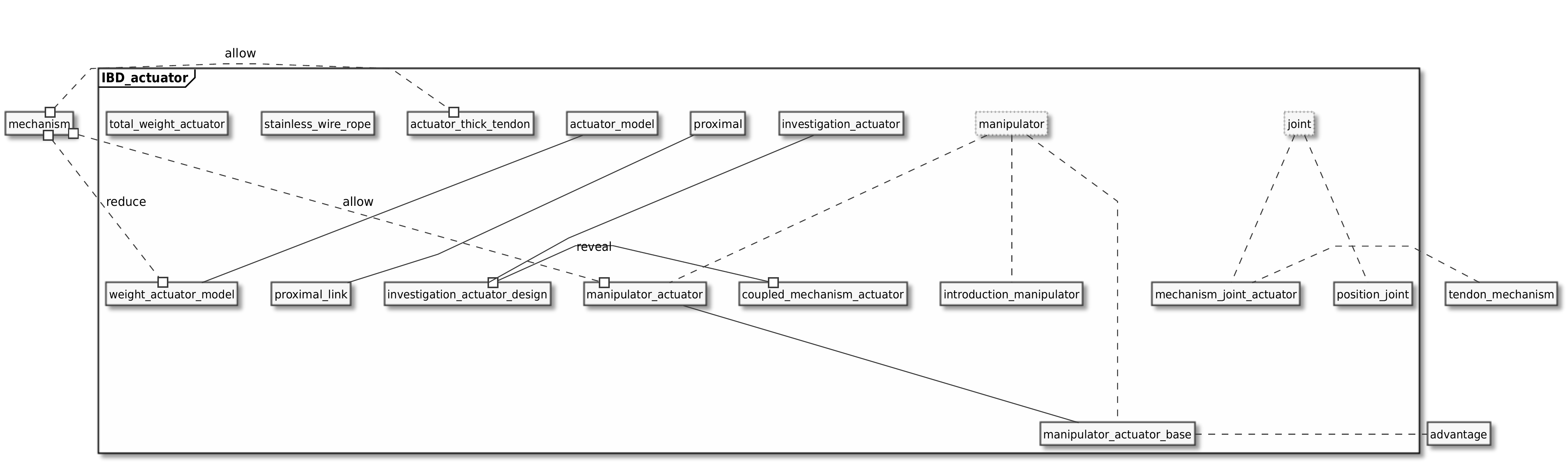}
    \end{subfigure}
    
    \begin{subfigure}[b]{\columnwidth}
    \centering
    \includegraphics[angle=270,width=3cm]{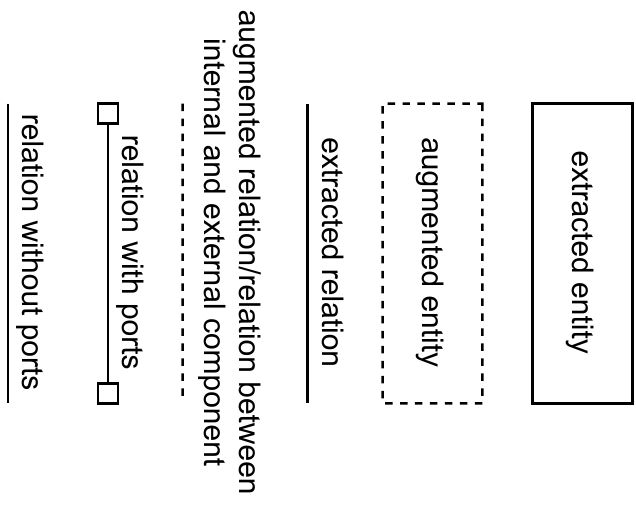}
    \end{subfigure}

\caption{IBD diagram for research paper dataset, selected phrase = `actuator'}
\label{figure:ibd}

\end{figure}

The IBD diagram specifies the connection between different sub-blocks of a given parent block. An example of the IBD diagram is illustrated in Figure \ref{figure:ibd}. It is assumed that two blocks are connected with ports if an action phrase exists in the relationship whose subject and object phrases define these two blocks. The presence of action phrases and corresponding ports is useful to show the interaction between different blocks, in addition to their hierarchical relationship described by the BDD diagram. Additionally, it can be seen that there are more blocks in the IBD than in the BDD diagram, because the IBD diagram uses a different algorithm to identify sub-blocks. This acts as a redundancy measure for the identification of sub-blocks that are not captured by the BDD algorithm, to help generate a more complete overview of the system. Additionally, reference relationships with blocks that are outside the user-specified block are also included in the IBD to highlight potential interactions and communications with external blocks. 

\subsection{REQ diagram generation}

\begin{figure}[p]
    \centering
    \begin{subfigure}[b]{\columnwidth}
    \centering
    \includegraphics[angle=270,width=4.5cm]{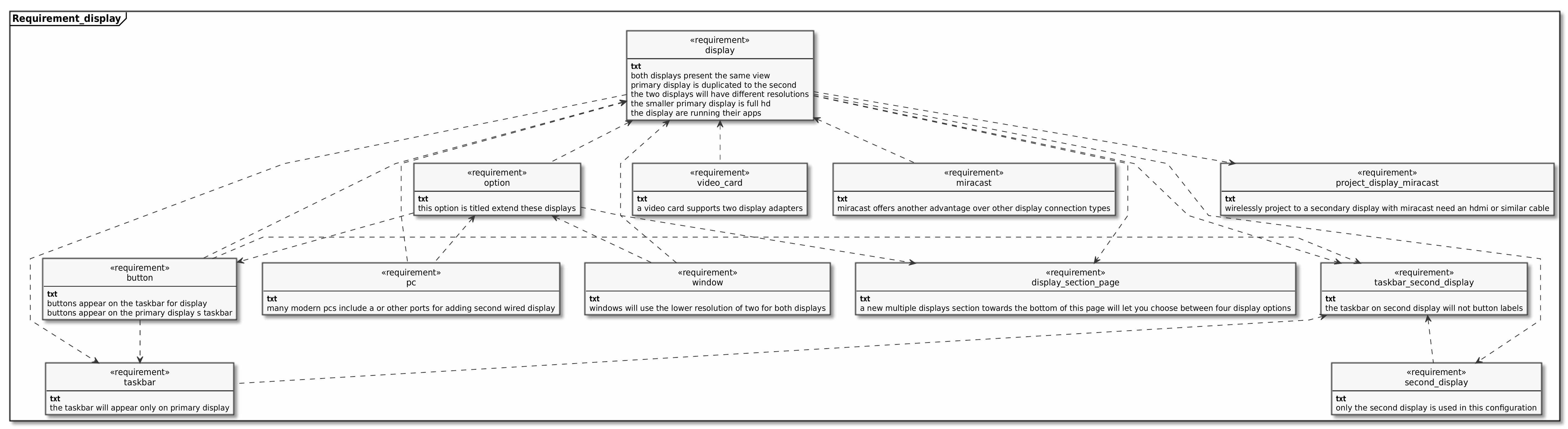}
    \end{subfigure}
    
    \begin{subfigure}[b]{\columnwidth}
    \centering
    \includegraphics[angle=270,width=0.8cm]{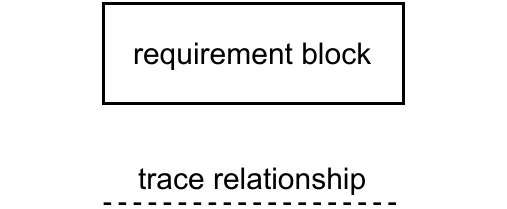}
    \end{subfigure}

\caption{REQ diagram for Windows manual dataset, selected phrase = `display'}
\label{figure:req}

\end{figure}

The REQ diagram specifies the requirements satisfied by the blocks and the relationships between different requirements. An example of the REQ diagram is illustrated in Figure \ref{figure:req}. It can be seen that some of the requirements share the same name and. are satisfied by the same block, implying that the algorithm has successfully clustered them together. Additionally, most of the requirements are connected to each other via augmented trace relationships. These can serve as candidates for more specific types of relationships for the user. The relationships also help create a hierarchical structure in the requirement diagram, where the most connected requirement is assumed to be placed higher in the hierarchy. Moreover, because the requirements are derived from relationships extracted from individual sentences, they are mostly functional requirements. The extraction of non-functional requirements is out of the scope of this study. By presenting the extracted relationships as requirements to the user, the algorithm is also able to provide more sentence context to help the users understand the phrases and relationships presented in BDD and IBD diagrams. 

From the generated figures, it can be seen that the algorithm is able to generate structured SysML diagrams that can potentially aid engineers in designing and architecting complex systems, alongside any existing diagrams and documentations they already possess. By defining the steps and procedures in extracting key phrases and classifying relationships, the approach provides a standardised, and arguably objective starting point for the user to understand and design different systems. Additionally, the approach also provides a degree of versatility in allowing the user to freely choose and mix the textual materials to upload, and to adjust the multiple parameters to achieve the desired specificity. 

\newpage
\newgeometry{top=1cm}

\restoregeometry

\newpage
\newgeometry{top=1cm}
\restoregeometry

\subsection{Time and space complexity of the SysML diagrams generation}
\textcolor{black}{The time taken for the automatic generation of SysML diagrams depends on the length of the document and on the computational resources available. For example, in the patent documents with around 9000 words and using a laptop machine with 8 GB RAM, the generation of SysML diagrams took around 5 minutes. However, the majority of the time was spent on extracting relationships using the OpenIE tool, which only needs to be done once for each document, and which does not require human involvement. In deployment, once the relationships are extracted (and saved to separate files), the time required to generate SysML diagrams is less than 1 minute, enabling the user to adjust the hyperparameters and visualise updated SysML outputs quickly.}

\textcolor{black}{Additionally, in deployment, the operational space required for generating the SysML diagrams is also minimal, as the intermediate files are all text documents, and the resolution of the generated SysML images can be adjusted according to the user’s need. The space required is less than 9 GB in total. However, most of the space needed is from the OpenIE toolbox and language models (8.76GB), which can live on a server and not necessarily on the user’s local machine. 
}

\section{Discussion}
\textcolor{black}{The work presented addresses the three research questions set in subsection \ref{subsection:researchq}  by yielding an approach that is}
\begin{itemize}
\item \textcolor{black}{Flexible and open-domain. By using techniques from open information extraction, it does not require the input document to follow specific syntactic rules or to belong to a specific domain. Instead, it is able to extract the required information for plotting from free, unstructured natural language text.}
\item \textcolor{black}{Automated: Takes natural language text as input and directly generates different types of SysML diagrams with no human intervention or intermediate modelling required. }
\item \textcolor{black}{Parameter-based: The proposed approach has minimal reliance on heuristic rules and predefined patterns to identify hierarchies. Instead, the approach parameterises the extraction of key entities and relationships and allows the user to adjust the degree of specificity using a set of parameters. }
\end{itemize}

\textcolor{black}{From the results presented in section \ref{section:results} it is possible to state that according to both quantitative and qualitative criteria the proposed approach is capable of outputting reasonable SysML diagrams, similar enough to those that could emerge from an engineer-made first draft. It is difficult to directly compare the results to similar work since, to the best of our knowledge, no attempts have been made, so far, to automatically generate SysML diagrams from natural text documents. Nevertheless, we regard this work as an encouraging first step towards enabling further and better automation of other systems engineering processes.} 

\textcolor{black}{Some extensions of the current work could include:}
\begin{itemize}
    \item \textcolor{black}{Development of an interactive tool from which engineers can correct and provide learning feedback to the algorithm as to what constitutes a good result or not. Such a tool could also allow for more explicit calibration and experimenting by the user of the various parameters that need tuning in order for the results to achieve sufficient quality;}
    \item \textcolor{black}{Quantification of the notion of \textit{completeness and consistency} in some meaningful way to allow comparison against human-designed diagrams;}
    \item \textcolor{black}{Adaptation of the method to other types of system modeling languages (e.g. Object Process Methodology (OPM)) \cite{dori2011object};}
    \item \textcolor{black}{Adaptation of the method to include Named Entity Recognition functionality.}
\end{itemize} 
\textcolor{black}{
One other aspect that is worth exploring (and that would perhaps call for an experimental setup including actual engineers working on a given task) would be to test how much of a detrimental impact could the reliance on automation have on the engineers' capacity to acquire and, most importantly, structure technical knowledge in the minds. As often happens when a previously manual task is automated, there is a risk of progressively deteriorating the level of preparation of the involved operators.}

\section{Conclusions}
\label{section:conclusions}
In this study, an approach to automatically generate SysML diagrams end-to-end directly from unstructured natural language text was proposed. This approach consists of six steps, and it leverages open-access tools such as NLTK, OpenIE, and PlantUML. \textcolor{black}{One of the key strengths of the proposed approach is} the combined use of Natural Language Processing techniques and the WordNet (available in NLTK) with the OpenIE toolbox. \textcolor{black}{This enabled the automatic} extraction of key textual relationships and phrases, and \textcolor{black}{for the first time,} the mapping and augmentation of these phrases and relationships to SysML model elements. The results are successfully leveraged to plot different structured SysML diagrams, by using the PlantUML tool. This approach requires five hyperparameters to be specified by the user, and it is open-domain. The versatility of the approach was demonstrated through the use of six case studies from different domains and using different writing styles. The results obtained with the proposed approach were validated against manually extracted results, which achieved high recall and precision in key phrase extraction, high accuracy in relationship classification, and success in generation of high-quality SysML diagrams.  

This approach would benefit systems engineers in gaining a detailed graphical overview of the system dependencies at the initial design phase, since it provides a standardised, comprehensive and automated starting point from textual resources (e.g., specifications, manuals, technical reports, maintenance reports). This will support systems engineers in understanding complex systems and in quickly assessing and designing the SysML diagrams for these systems. As far as the authors are aware, this study is the first attempt at directly generating SysML diagrams from natural language text.

\section*{Acknowledgement}
The authors would like to thank Prof. Janet Pierrehumbert (University of Oxford) and Prof. Youssef Marzouk (MIT) for the helpful comments and insights at the early stages of this project. Part of this work was supported by the Department of Engineering Science at the University of Oxford through the Engineering Undergraduate Research Opportunities (EUROP) in 2019 (recipient SZ, supervisors AC and AS), and by Balliol College, University of Oxford through the Career Development Fellowship in Engineering Science of AC and a research bursary to SZ.

\newpage
\section{APPENDIX}
\appendix
\section{Table of symbols}
\begin{longtable}{p{.15\textwidth}|p{.75\textwidth}} 
Symbol & Meaning \\ \hline
$w$ & Tf-idf weighting \\ \hline
$t$ & Word \\ \hline
$d$ & Document \\ \hline
$c$ & Corpus of documents \\ \hline
$w_{t,d}$ & Tf-idf weighting of a word in a document\\ \hline
$\textup{tf}_{t,d}$ & Term frequency of a word in a document \\ \hline
$\textup{idf}_{t,c}$ & Inverse document frequency of a word in a corpus of documents \\ \hline
$N_c$ & Total number of documents in a corpus \\ \hline
$\textup{df}_{t,c}$ & Document frequency, the number of documents in the corpus that contain a particular word \\ \hline
$\sigma_{tf-idf}$ &  Manually specified tf-idf weighting threshold for selection of key nouns \\ \hline
$\sigma_{{relationship}}$ & Manually specified confidence threshold to preliminarily filter extracted relationships \\ \hline
$r$ & Textual relationships \\ \hline
$p$ & Phrases \\ \hline
$p_{1,2,3,...}$ & Ordered phrases in a textual relationship \\ \hline
$N_r$ & Number of phrases in an extracted relationship \\ \hline
$t_{1,2,3,...}$ & Ordered words in a phrase \\ \hline
$N_p$ & Number of words in a phrase \\ \hline
$L_{phrase}$ & Manually specified maximum number of nouns in a key phrase \\ \hline
$count_{p,d}$ & Frequency of a phrase, referring to the total count of a phrase in a document \\ \hline
$h'$ & WordNet depth, the semantic depth assigned by WordNet to an individual word given a context \\ \hline
$h$ & WordNet score, 1-complement of the normalised WordNet depth of a word  \\ \hline
$\lambda_{p,d}$ & Score of a candidate key phrase given a document \\ \hline
$\sigma_{p}$ & Manually specified score threshold for selection of key phrases from candidate key phrases \\ \hline
$B$ & SysML block \\ \hline
$R$ & SysML relationship between blocks \\ \hline
$B_{p}$ & SysML block defined by a phrase \\ \hline
$\sigma_{rel-difference}$ & Manually specified score threshold for classifying between composite and reference relationships \\ \hline
$P$ & Top level phrases, the set of key phrases whose corresponding blocks do not form a sub-block of another \\ \hline
$\gamma$ & Score of a word, which is a sum of tf-idf weighting and WordNet score \\ \hline
$\gamma_{t,d}$ & Score of a word in a document \\ \hline
$\Gamma_{p,d}$ & List of the scores of words in a phrase of a document \\ \hline
$p_{abstract}$ & Abstracted phrase \\ \hline
\caption{The definition of symbols used in the paper}
\label{table:symboltable}
\end{longtable}

\bibliography{mybibfile}

\begin{thebibliography}{10}
\expandafter\ifx\csname url\endcsname\relax
  \def\url#1{\texttt{#1}}\fi
\expandafter\ifx\csname urlprefix\endcsname\relax\def\urlprefix{URL }\fi
\expandafter\ifx\csname href\endcsname\relax
  \def\href#1#2{#2} \def\path#1{#1}\fi

\bibitem{kossiakoff2011systems}
A.~Kossiakoff, W.~Sweet, S.~Seymour, S.~Biemer, Systems Engineering Principles
  and Practice, 2nd Edition, John Wiley \& Sons, Ltd, 2011.
\newblock \href {http://dx.doi.org/doi.org/10.1002/9781118001028}
  {\path{doi:doi.org/10.1002/9781118001028}}.

\bibitem{friedenthal14practical}
S.~Friedenthal, A.~Moore, R.~Steiner, A Practical Guide to SysML, Third
  Edition: The Systems Modeling Language, 3rd Edition, Morgan Kaufmann
  Publishers Inc., San Francisco, CA, USA, 2014.
\newblock \href {http://dx.doi.org/https://doi.org/10.1016/C2013-0-14457-1}
  {\path{doi:https://doi.org/10.1016/C2013-0-14457-1}}.

\bibitem{hart2015introduction}
L.~E. Hart, Introduction to model-based system engineering (mbse) and sysml,
  in: Delaware Valley INCOSE Chapter Meeting, Vol.~30, Ramblewood Country Club
  Mount Laurel, New Jersey, 2015.

\bibitem{delligatti2013sysml}
L.~Delligatti, SysML Distilled: A Brief Guide to the Systems Modeling Language,
  1st Edition, Addison-Wesley Professional, 2013.

\bibitem{huang2007system}
E.~Huang, R.~Ramamurthy, L.~F. McGinnis, System and simulation modeling using
  sysml, in: 2007 Winter Simulation Conference, IEEE, 2007, pp. 796--803, {ISBN
  = 978-1-4244-1305-8.} ISSN = 1558-4305. INSPEC Accession Number = 9847802.
\newblock \href {http://dx.doi.org/10.1109/WSC.2007.4419675}
  {\path{doi:10.1109/WSC.2007.4419675}}.

\bibitem{dori2004smart}
D.~Dori, N.~Korda, A.~Soffer, S.~Cohen, Smart: System model acquisition from
  requirements text, in: J.~Desel, B.~Pernici, M.~Weske (Eds.), Business
  Process Management, Springer Berlin Heidelberg, Berlin, Heidelberg, 2004, pp.
  179--194, {Lecture Notes in Computer Science, vol 3080.}
\newblock \href {http://dx.doi.org/10.1007/978-3-540-25970-1_12}
  {\path{doi:10.1007/978-3-540-25970-1_12}}.

\bibitem{sawyer2005shallow}
P.~Sawyer, P.~Rayson, K.~Cosh, Shallow knowledge as an aid to deep
  understanding in early phase requirements engineering, IEEE Transactions on
  Software Engineering 31~(11) (2005) 969--981, electronic ISSN: 1939-3520.
  INSPEC Accession Number: 8727391.
\newblock \href {http://dx.doi.org/10.1109/TSE.2005.129}
  {\path{doi:10.1109/TSE.2005.129}}.

\bibitem{arellano2015frameworks}
A.~Arellano, E.~Zontek-Carney, M.~A. Austin, Frameworks for natural language
  processing of textual requirements, International Journal On Advances in
  Systems and Measurements 8 (2015) 230--240.

\bibitem{omg19omg}
{Object Management Group}, \href{https://www.omg.org/spec/SysML/1.6}{{OMG
  System Modeling Language Version 1.6}}, Standard, {Object Management Group
  ({OMG})} (Dec. 2019).
\newline\urlprefix\url{https://www.omg.org/spec/SysML/1.6}

\bibitem{friedenthal08omg}
S.~Friedenthal, A.~Moore, R.~Steiner, {OMG Systems Modeling Language (OMG
  SysML) Tutorial}, INCOSE International Symposium 18 (2008) 1731--1862.
\newblock \href {http://dx.doi.org/10.1002/j.2334-5837.2008.tb00914.x}
  {\path{doi:10.1002/j.2334-5837.2008.tb00914.x}}.

\bibitem{hause06thesysml}
M.~Hause, {The SysML Modelling Language}, in: Fifteenth European Systems
  Engineering Conference, 2006.

\bibitem{zhao2021natural}
L.~Zhao, W.~Alhoshan, A.~Ferrari, K.~J. Letsholo, M.~A. Ajagbe, E.-V. Chioasca,
  R.~T. Batista-Navarro, Natural language processing for requirements
  engineering: A systematic mapping study, ACM Comput. Surv. 54~(3).
\newblock \href {http://dx.doi.org/10.1145/3444689}
  {\path{doi:10.1145/3444689}}.

\bibitem{bakar2016extracting}
N.~H. Bakar, Z.~M. Kasirun, N.~Salleh, H.~A. Jalab, Extracting features from
  online software reviews to aid requirements reuse, Applied Soft Computing 49
  (2016) 1297--1315.
\newblock \href {http://dx.doi.org/https://doi.org/10.1016/j.asoc.2016.07.048}
  {\path{doi:https://doi.org/10.1016/j.asoc.2016.07.048}}.

\bibitem{robeer2016automated}
M.~Robeer, G.~Lucassen, J.~M. E.~M. van~der Werf, F.~Dalpiaz, S.~Brinkkemper,
  Automated extraction of conceptual models from user stories via nlp, in: 2016
  IEEE 24th International Requirements Engineering Conference (RE), 2016, pp.
  196--205, {ISBN:978-1-5090-4121-3}. ISSN: 2332-6441. INSPEC Accession Number:
  16519370.
\newblock \href {http://dx.doi.org/10.1109/RE.2016.40}
  {\path{doi:10.1109/RE.2016.40}}.

\bibitem{johann2017safe}
T.~Johann, C.~Stanik, A.~M. Alizadeh~B., W.~Maalej, Safe: A simple approach for
  feature extraction from app descriptions and app reviews, in: 2017 IEEE 25th
  International Requirements Engineering Conference (RE), 2017, pp. 21--30,
  {ISBN: 978-1-5386-3191-1. ISSN: 2332-6441. INSPEC Accession Number:
  17207660}.
\newblock \href {http://dx.doi.org/10.1109/RE.2017.71}
  {\path{doi:10.1109/RE.2017.71}}.

\bibitem{abad2019supporting}
Z.~Shakeri Hossein~Abad, V.~Gervasi, D.~Zowghi, B.~H.~Far, Supporting analysts
  by dynamic extraction and classification of requirements-related knowledge,
  in: 2019 IEEE/ACM 41st International Conference on Software Engineering
  (ICSE), 2019, pp. 442--453, {ISBN:978-1-7281-0869-8. ISSN: 1558-1225. INSPEC
  Accession Number: 18938395}.
\newblock \href {http://dx.doi.org/10.1109/ICSE.2019.00057}
  {\path{doi:10.1109/ICSE.2019.00057}}.

\bibitem{korner2010semantic}
S.~J. K{\"o}rner, M.~Landh{\"a}u{\ss}er, Semantic enriching of natural language
  texts with automatic thematic role annotation, in: C.~J. Hopfe, Y.~Rezgui,
  E.~M{\'e}tais, A.~Preece, H.~Li (Eds.), Natural Language Processing and
  Information Systems, Springer Berlin Heidelberg, Berlin, Heidelberg, 2010,
  pp. 92--99, {Lecture Notes in Computer Science, vol 6177.}
\newblock \href {http://dx.doi.org/10.1007/978-3-642-13881-2_9}
  {\path{doi:10.1007/978-3-642-13881-2_9}}.

\bibitem{hayes2003improving}
J.~Hayes, A.~Dekhtyar, J.~Osborne, Improving requirements tracing via
  information retrieval, in: Proceedings. 11th IEEE International Requirements
  Engineering Conference, 2003., 2003, pp. 138--147, {ISBN:0-7695-1980-6. ISSN:
  1090-705X. INSPEC Accession Number: 7913997}.
\newblock \href {http://dx.doi.org/10.1109/ICRE.2003.1232745}
  {\path{doi:10.1109/ICRE.2003.1232745}}.

\bibitem{casamayor2010identification}
A.~Casamayor, D.~Godoy, M.~Campo, Identification of non-functional requirements
  in textual specifications: A semi-supervised learning approach, Information
  and Software Technology 52~(4) (2010) 436--445.
\newblock \href
  {http://dx.doi.org/https://doi.org/10.1016/j.infsof.2009.10.010}
  {\path{doi:https://doi.org/10.1016/j.infsof.2009.10.010}}.

\bibitem{rodriguez2019efficient}
D.~V. Rodriguez, D.~L. Carver, A.~Mahmoud, An efficient wikipedia-based
  approach for better understanding of natural language text related to user
  requirements, in: 2018 IEEE Aerospace Conference, 2018, pp. 1--16,
  {ISBN:978-1-5386-2014-4}.
\newblock \href {http://dx.doi.org/10.1109/AERO.2018.8396645}
  {\path{doi:10.1109/AERO.2018.8396645}}.

\bibitem{ferrari2013mining}
A.~Ferrari, G.~O. Spagnolo, F.~Dell'Orletta, Mining commonalities and
  variabilities from natural language documents, in: Proceedings of the 17th
  International Software Product Line Conference, SPLC '13, Association for
  Computing Machinery, New York, NY, USA, 2013, p. 116–120.
\newblock \href {http://dx.doi.org/10.1145/2491627.2491634}
  {\path{doi:10.1145/2491627.2491634}}.

\bibitem{thakur2016identifying}
J.~S. Thakur, A.~Gupta, Identifying domain elements from textual
  specifications, in: Proceedings of the 31st IEEE/ACM International Conference
  on Automated Software Engineering, ASE 2016, Association for Computing
  Machinery, New York, NY, USA, 2016, p. 566–577, {INSPEC Accession Number:
  16358075}.
\newblock \href {http://dx.doi.org/10.1145/2970276.2970323}
  {\path{doi:10.1145/2970276.2970323}}.

\bibitem{ferrari2019nlp}
A.~Ferrari, A.~Esuli, An nlp approach for cross-domain ambiguity detection in
  requirements engineering, Automated Software Engineering 26 (2019) 559--598.
\newblock \href {http://dx.doi.org/10.1007/s10515-019-00261-7}
  {\path{doi:10.1007/s10515-019-00261-7}}.

\bibitem{carvalho2013test}
G.~Carvalho, D.~Falc\~{a}o, F.~Barros, A.~Sampaio, A.~Mota, L.~Motta,
  M.~Blackburn, Test case generation from natural language requirements based
  on scr specifications, in: Proceedings of the 28th Annual ACM Symposium on
  Applied Computing, SAC '13, Association for Computing Machinery, New York,
  NY, USA, 2013, p. 1217–1222.
\newblock \href {http://dx.doi.org/10.1145/2480362.2480591}
  {\path{doi:10.1145/2480362.2480591}}.

\bibitem{silva2016test}
B.~C.~F. Silva, G.~Carvalho, A.~Sampaio, Test case generation from natural
  language requirements using cpn simulation, in: M.~Corn{\'e}lio, B.~Roscoe
  (Eds.), Formal Methods: Foundations and Applications, Springer International
  Publishing, Cham, 2016, pp. 178--193.
\newblock \href {http://dx.doi.org/10.1007/978-3-319-29473-5_11}
  {\path{doi:10.1007/978-3-319-29473-5_11}}.

\bibitem{tiwari2019approach}
S.~Tiwari, D.~Ameta, A.~Banerjee, An approach to identify use case scenarios
  from textual requirements specification, in: Proceedings of the 12th
  Innovations on Software Engineering Conference (Formerly Known as India
  Software Engineering Conference), ISEC'19, Association for Computing
  Machinery, New York, NY, USA, 2019, pp. 1--11.
\newblock \href {http://dx.doi.org/10.1145/3299771.3299774}
  {\path{doi:10.1145/3299771.3299774}}.

\bibitem{ferrari2018detecting}
A.~Ferrari, G.~Gori, B.~Rosadini, I.~Trotta, S.~Bacherini, A.~Fantechi,
  S.~Gnesi, Detecting requirements defects with nlp patterns: An industrial
  experience in the railway domain, Empirical Softw. Engg. 23~(6) (2018)
  3684–3733.
\newblock \href {http://dx.doi.org/10.1007/s10664-018-9596-7}
  {\path{doi:10.1007/s10664-018-9596-7}}.

\bibitem{loughran2006from}
N.~Loughran, A.~Sampaio, A.~Rashid, From requirements documents to feature
  models for aspect oriented product line implementation, in: J.-M. Bruel
  (Ed.), Satellite Events at the MoDELS 2005 Conference, Springer Berlin
  Heidelberg, Berlin, Heidelberg, 2006, pp. 262--271, {Lecture Notes in
  Computer Science, vol 3844.}
\newblock \href {http://dx.doi.org/doi.org/10.1007/11663430_27}
  {\path{doi:doi.org/10.1007/11663430_27}}.

\bibitem{sreekumar2018extracting}
A.~Sree-Kumar, E.~Planas, R.~Claris\'{o}, Extracting software product line
  feature models from natural language specifications, in: Proceedings of the
  22nd International Systems and Software Product Line Conference - Volume 1,
  SPLC '18, Association for Computing Machinery, New York, NY, USA, 2018, p.
  43–53.
\newblock \href {http://dx.doi.org/10.1145/3233027.3233029}
  {\path{doi:10.1145/3233027.3233029}}.

\bibitem{al2009natural}
L.~A. Al-Safadi, Natural language processing for conceptual modeling,
  International Journal of Digital Content Technology and its Applications
  3~(3) (2009) 47--59.

\bibitem{casagrande2014nlp}
E.~Casagrande, S.~Woldeamlak, W.~L. Woon, H.~H. Zeineldin, D.~Svetinovic,
  Nlp-kaos for systems goal elicitation: Smart metering system case study, IEEE
  Transactions on Software Engineering 40~(10) (2014) 941--956, {ISSN:
  1939-3520. INSPEC Accession Number: 14652000}.
\newblock \href {http://dx.doi.org/10.1109/TSE.2014.2339811}
  {\path{doi:10.1109/TSE.2014.2339811}}.

\bibitem{thayasivam2011automatically}
U.~Thayasivam, K.~Verma, A.~Kass, R.~Vasquez,
  \href{https://ojs.aaai.org/index.php/AAAI/article/view/18863}{Automatically
  mapping natural language requirements to domain-specific process models},
  Proceedings of the AAAI Conference on Artificial Intelligence 25~(2) (2011)
  1695--1700.
\newline\urlprefix\url{https://ojs.aaai.org/index.php/AAAI/article/view/18863}

\bibitem{nguyen2015rule}
T.~H. Nguyen, J.~Grundy, M.~Almorsy, Rule-based extraction of goal-use case
  models from text, in: Proceedings of the 2015 10th Joint Meeting on
  Foundations of Software Engineering, ESEC/FSE 2015, Association for Computing
  Machinery, New York, NY, USA, 2015, p. 591–601.
\newblock \href {http://dx.doi.org/10.1145/2786805.2786876}
  {\path{doi:10.1145/2786805.2786876}}.

\bibitem{chen2009automatic}
L.~Chen, Y.~Zeng, Automatic generation of uml diagrams from product
  requirements described by natural language, Vol. Volume 2: 29th Computers and
  Information in Engineering Conference, Parts A and B of International Design
  Engineering Technical Conferences and Computers and Information in
  Engineering Conference, 2009, pp. 779--786.
\newblock \href {http://dx.doi.org/10.1115/DETC2009-86514}
  {\path{doi:10.1115/DETC2009-86514}}.

\bibitem{zeng2008recursive}
Y.~Zeng, Recursive object model (rom)-modelling of linguistic information in
  engineering design, Computers in Industry 59~(6) (2008) 612–625.
\newblock \href {http://dx.doi.org/10.1016/j.compind.2008.03.002}
  {\path{doi:10.1016/j.compind.2008.03.002}}.

\bibitem{afreen2011generating}
H.~Afreen, I.~S. Bajwa, {Generating UML class models from SBVR software
  requirements specifications}, in: 23rd Benelux Conference on Artificial
  Intelligence (BNAIC 2011), Citeseer, 2011, pp. 23--32.

\bibitem{deeptimahanti2011semiautomatic}
D.~K. Deeptimahanti, R.~Sanyal, Semi-automatic generation of uml models from
  natural language requirements, in: Proceedings of the 4th India Software
  Engineering Conference, ISEC '11, Association for Computing Machinery, New
  York, NY, USA, 2011, p. 165–174.
\newblock \href {http://dx.doi.org/10.1145/1953355.1953378}
  {\path{doi:10.1145/1953355.1953378}}.

\bibitem{gruber1995toward}
T.~R. Gruber, Toward principles for the design of ontologies used for knowledge
  sharing, International journal of human-computer studies 43~(5-6) (1995)
  907--928.
\newblock \href {http://dx.doi.org/10.1006/ijhc.1995.1081}
  {\path{doi:10.1006/ijhc.1995.1081}}.

\bibitem{asim2018survey}
M.~N. Asim, M.~Wasim, M.~U.~G. Khan, W.~Mahmood, H.~M. Abbasi, {A survey of
  ontology learning techniques and applications}, Database 2018.
\newblock \href {http://dx.doi.org/10.1093/database/bay101}
  {\path{doi:10.1093/database/bay101}}.

\bibitem{cimiano2005text2onto}
P.~Cimiano, J.~V\"{o}lker, Text2onto: A framework for ontology learning and
  data-driven change discovery, in: Proceedings of the 10th International
  Conference on Natural Language Processing and Information Systems, NLDB'05,
  Springer-Verlag, Berlin, Heidelberg, 2005, p. 227–238.
\newblock \href {http://dx.doi.org/10.1007/11428817_21}
  {\path{doi:10.1007/11428817_21}}.

\bibitem{drymonas2010unsupervised}
E.~Drymonas, K.~Zervanou, E.~G.~M. Petrakis, Unsupervised ontology acquisition
  from plain texts: The ontogain system, in: Proceedings of the Natural
  Language Processing and Information Systems, and 15th International
  Conference on Applications of Natural Language to Information Systems,
  NLDB'10, Springer-Verlag, Berlin, Heidelberg, 2010, p. 277–287.
\newblock \href {http://dx.doi.org/10.1007/978-3-642-13881-2_29}
  {\path{doi:10.1007/978-3-642-13881-2_29}}.

\bibitem{velardi2013ontolearn}
P.~Velardi, S.~Faralli, R.~Navigli, {OntoLearn Reloaded: A Graph-Based
  Algorithm for Taxonomy Induction}, Computational Linguistics 39~(3) (2013)
  665--707.
\newblock \href {http://dx.doi.org/10.1162/COLI_a_00146}
  {\path{doi:10.1162/COLI_a_00146}}.

\bibitem{jiang2010crctol}
X.~Jiang, A.-H. Tan, {CRCTOL: A Semantic-Based Domain Ontology Learning
  System}, Journal of the American Society for Information Science and
  Technology 61~(1) (2010) 150–168.
\newblock \href {http://dx.doi.org/10.1002/asi.21231}
  {\path{doi:10.1002/asi.21231}}.

\bibitem{mejhedmkhinini2020combining}
M.~{Mejhed Mkhinini}, O.~Labbani-Narsis, C.~Nicolle, Combining uml and
  ontology: An exploratory survey, Computer Science Review 35 (2020) 100223.
\newblock \href
  {http://dx.doi.org/https://doi.org/10.1016/j.cosrev.2019.100223}
  {\path{doi:https://doi.org/10.1016/j.cosrev.2019.100223}}.

\bibitem{jurafsky2009speech}
D.~Jurafsky, J.~H. Martin, Speech and Language Processing (2nd Edition),
  Prentice-Hall, Inc., USA, 2009, {ISBN = 0131873210}.

\bibitem{kiss2006unsupervised}
T.~Kiss, J.~Strunk,
  \href{https://doi.org/10.1162/coli.2006.32.4.485}{{Unsupervised Multilingual
  Sentence Boundary Detection}}, Computational Linguistics 32~(4) (2006)
  485--525.
\newblock \href
  {http://arxiv.org/abs/https://direct.mit.edu/coli/article-pdf/32/4/485/1798345/coli.2006.32.4.485.pdf}
  {\path{arXiv:https://direct.mit.edu/coli/article-pdf/32/4/485/1798345/coli.2006.32.4.485.pdf}},
  \href {http://dx.doi.org/10.1162/coli.2006.32.4.485}
  {\path{doi:10.1162/coli.2006.32.4.485}}.
\newline\urlprefix\url{https://doi.org/10.1162/coli.2006.32.4.485}

\bibitem{bird2009natural}
S.~Bird, E.~Klein, E.~Loper, Natural language processing with Python: analyzing
  text with the natural language toolkit, O'Reilly Media, Inc., 2009.

\bibitem{luhn1957statistical}
H.~P. Luhn, A statistical approach to mechanized encoding and searching of
  literary information, IBM Journal of research and development 1~(4) (1957)
  309--317.

\bibitem{jones1972statistical}
K.~S. Jones, A statistical interpretation of term specificity and its
  application in retrieval, Journal of documentation.

\bibitem{mausam2016open}
M.~Mausam, Open information extraction systems and downstream applications, in:
  Proceedings of the twenty-fifth international joint conference on artificial
  intelligence, 2016, pp. 4074--4077.

\bibitem{christensen2011analysis}
J.~Christensen, Mausam, S.~Soderland, O.~Etzioni, An analysis of open
  information extraction based on semantic role labeling, in: Proceedings of
  the Sixth International Conference on Knowledge Capture, K-CAP '11,
  Association for Computing Machinery, New York, NY, USA, 2011, p. 113–120.
\newblock \href {http://dx.doi.org/10.1145/1999676.1999697}
  {\path{doi:10.1145/1999676.1999697}}.

\bibitem{pal2016demonyms}
H.~Pal, {Mausam}, Demonyms and compound relational nouns in nominal open {IE},
  in: Proceedings of the 5th Workshop on Automated Knowledge Base Construction,
  Association for Computational Linguistics, San Diego, CA, 2016, pp. 35--39.
\newblock \href {http://dx.doi.org/10.18653/v1/W16-1307}
  {\path{doi:10.18653/v1/W16-1307}}.

\bibitem{saha2017bootstrapping}
S.~Saha, H.~Pal, {Mausam},
  \href{https://aclanthology.org/P17-2050}{Bootstrapping for numerical open
  {IE}}, in: Proceedings of the 55th Annual Meeting of the Association for
  Computational Linguistics (Volume 2: Short Papers), Association for
  Computational Linguistics, Vancouver, Canada, 2017, pp. 317--323.
\newblock \href {http://dx.doi.org/10.18653/v1/P17-2050}
  {\path{doi:10.18653/v1/P17-2050}}.
\newline\urlprefix\url{https://aclanthology.org/P17-2050}

\bibitem{saha2018open}
S.~Saha, {Mausam}, \href{https://aclanthology.org/C18-1194}{Open information
  extraction from conjunctive sentences}, in: Proceedings of the 27th
  International Conference on Computational Linguistics, Association for
  Computational Linguistics, Santa Fe, New Mexico, USA, 2018, pp. 2288--2299.
\newline\urlprefix\url{https://aclanthology.org/C18-1194}

\bibitem{punyakanok2008importance}
V.~Punyakanok, D.~Roth, W.-t. Yih, The importance of syntactic parsing and
  inference in semantic role labeling, Computational Linguistics 34~(2) (2008)
  257--287.

\bibitem{johansson2008effect}
R.~Johansson, P.~Nugues, The effect of syntactic representation on semantic
  role labeling, in: Proceedings of the 22nd International Conference on
  Computational Linguistics - Volume 1, COLING '08, Association for
  Computational Linguistics, USA, 2008, p. 393–400.

\bibitem{pauls2011faster}
A.~Pauls, D.~Klein, Faster and smaller <i>n</i>-gram language models, in:
  Proceedings of the 49th Annual Meeting of the Association for Computational
  Linguistics: Human Language Technologies - Volume 1, HLT '11, Association for
  Computational Linguistics, USA, 2011, p. 258–267.

\bibitem{fellbaum98wordnet}
C.~Fellbaum, WordNet: An Electronic Lexical Database, Bradford Books, 1998.

\bibitem{WordNet}
{Princeton University}, \href{https://wordnet.princeton.edu/}{{About WordNet}}
  (2010).
\newline\urlprefix\url{https://wordnet.princeton.edu/}

\bibitem{lesk1986automatic}
M.~Lesk, Automatic sense disambiguation using machine readable dictionaries:
  How to tell a pine cone from an ice cream cone, in: Proceedings of the 5th
  Annual International Conference on Systems Documentation, SIGDOC '86,
  Association for Computing Machinery, New York, NY, USA, 1986, p. 24–26.
\newblock \href {http://dx.doi.org/10.1145/318723.318728}
  {\path{doi:10.1145/318723.318728}}.

\bibitem{plantuml}
{PlantUML Team}, \href{http://plantuml.com/}{{PlantUML}} (Oct 2021).
\newline\urlprefix\url{http://plantuml.com/}

\bibitem{plantumlguide}
{PlantUML Team}, \href{http://plantuml.com/guide}{{Drawing UML with PlantUML:
  PlantUML Language Reference Guide}} (2021).
\newline\urlprefix\url{http://plantuml.com/guide}

\bibitem{ellson2002graphviz}
J.~Ellson, E.~Gansner, L.~Koutsofios, S.~C. North, G.~Woodhull, Graphviz---
  open source graph drawing tools, in: P.~Mutzel, M.~J{\"u}nger, S.~Leipert
  (Eds.), Graph Drawing, Springer Berlin Heidelberg, Berlin, Heidelberg, 2002,
  pp. 483--484.

\bibitem{epopatent}
{European Patent Office},
  \href{https://www.epo.org/searching-for-patents/data/bulk-data-sets/text-analytics.html}{{EP
  full-text data for text analytics}}, licensed under Creative Commons
  Attribution 4.0 International Public License
  https://creativecommons.org/licenses/by/4.0/ (2021).
\newline\urlprefix\url{https://www.epo.org/searching-for-patents/data/bulk-data-sets/text-analytics.html}

\bibitem{windows10fieldguide}
P.~Thurrott, R.~Riveram, M.~McClean, Windows 10 Field Guide, Leanpub, 2021,
  [accessed 25-Mar-2021].

\bibitem{ukgovreport}
Arup, S.~Broadberry, T.~Leunig, J.~R. Bryson, J.~Clark, R.~Mulhall, H.-J.
  Chang, A.~Andreoni, M.~L. Kuan, S.~Deakin, et~al.,
  \href{https://www.gov.uk/government/collections/future-of-manufacturing#project-report}{{Future
  of manufacturing (2013)}}, [accessed 25-Mar-2021].
\newline\urlprefix\url{https://www.gov.uk/government/collections/future-of-manufacturing#project-report}

\bibitem{appstore}
finnqiao, \href{https://github.com/finnqiao/apple_appstore}{{Apple Appstore
  Descriptions}}, [accessed 25-Mar-2021].
\newline\urlprefix\url{https://github.com/finnqiao/apple_appstore}

\bibitem{ieeexplore}
IEEE, \href{https://ieeexplore.ieee.org/}{{IEEE Xplore}}, [accessed
  19-Mar-2021].
\newline\urlprefix\url{https://ieeexplore.ieee.org/}

\bibitem{wikipedia}
{Wikipedia contributors},
  \href{https://en.wikipedia.org/w/index.php?title=Member_states_of_the_United_Nations&oldid=1052617810}{Member
  states of the united nations --- {Wikipedia}{,} the free encyclopedia},
  [accessed 13-November-2020] (2021).
\newline\urlprefix\url{https://en.wikipedia.org/w/index.php?title=Member_states_of_the_United_Nations&oldid=1052617810}

\bibitem{hariri2013supporting}
N.~Hariri, C.~Castro-Herrera, M.~Mirakhorli, J.~Cleland-Huang, B.~Mobasher,
  Supporting domain analysis through mining and recommending features from
  online product listings, IEEE Transactions on Software Engineering 39~(12)
  (2013) 1736--1752.

\bibitem{dori2011object}
D.~Dori, Object-process methodology, in: Encyclopedia of Knowledge Management,
  Second Edition, IGI Global, 2011, pp. 1208--1220.

\end{thebibliography}

\end{document}